\documentclass[lettersize,journal]{IEEEtran}
\usepackage{amsmath,amsfonts}
\usepackage{algorithmic}
\usepackage{algorithm}
\usepackage{array}
\usepackage[caption=false,font=footnotesize,labelfont=rm,textfont=rm]{subfig}
\usepackage{textcomp}
\usepackage{makecell}
\usepackage{stfloats}
\usepackage{url}
\usepackage{verbatim}
\usepackage{graphicx}
\usepackage{cite}
\usepackage{colortbl} 
\usepackage{hyperref} 
\usepackage{bigstrut}
\usepackage{caption}
\usepackage{booktabs,makecell, multirow, tabularx}
\hypersetup{
    colorlinks=true, 
    linkcolor=blue, 
    citecolor=green, 
    urlcolor=blue 
}
\begin{document}

\title{

HyPSAM: Hybrid Prompt-driven Segment Anything Model for RGB-Thermal Salient Object Detection
}

\author{Ruichao Hou, \IEEEmembership{Member, IEEE}, Xingyuan Li, Tongwei Ren, \IEEEmembership{Member, IEEE}, Dongming Zhou, Gangshan Wu, \IEEEmembership{Member, IEEE} and Jinde Cao, \IEEEmembership{Fellow, IEEE}

\thanks{This work was supported by the National Natural Science Foundation of China (62072232), the Key R\&D Project of Jiangsu Province (BE2022138), the Fundamental Research Funds for the Central Universities (021714380026), the program B for Outstanding Ph.D. candidate of Nanjing University, and the Collaborative Innovation Center of Novel Software Technology and Industrialization. \emph{(Corresponding authors: Tongwei Ren)} }

\thanks{Ruichao Hou, Xingyuan Li, Tongwei Ren, Gangshan Wu are with the State Key Laboratory for Novel Software Technology, Nanjing University, Nanjing 210008, China (e-mail: rchou@nju.edu.cn; lixy@smail.nju.edu.cn; rentw@nju.edu.cn; gswu@nju.edu.cn).}
\thanks{Dongming Zhou is with the School of Information Science and Engineering, Yunnan University, Kunming 650091, China (e-mail: zhoudm@ynu.edu.cn).}
\thanks{Jinde Cao is with the School of Mathematics, Southeast University, Nanjing 210096, China (e-mail: jdcao@seu.edu.cn).}

}

\markboth{Submitted to IEEE Journals/Transactions}%
{Shell \MakeLowercase{\textit{\emph{et al.}}}: A Sample Article Using IEEEtran.cls for IEEE Journals}


\maketitle

\begin{abstract}
RGB-thermal salient object detection (RGB-T SOD) aims to identify prominent objects by integrating complementary information from RGB and thermal modalities.
However, learning the precise boundaries and complete objects remains challenging due to the intrinsic insufficient feature fusion and the extrinsic limitations of data scarcity.
In this paper, we propose a novel hybrid prompt-driven segment anything model (HyPSAM), which leverages the zero-shot generalization capabilities of the segment anything model (SAM) for RGB-T SOD. 
Specifically, we first propose a dynamic fusion network (DFNet) that generates high-quality initial saliency maps as visual prompts. DFNet employs dynamic convolution and multi-branch decoding to facilitate adaptive cross-modality interaction, overcoming the limitations of fixed-parameter kernels and enhancing multi-modal feature representation.
Moreover, we propose a plug-and-play refinement network (P2RNet) 
which serves as a general optimization strategy to guide SAM in refining saliency maps by using hybrid prompts. 
The text prompt ensures reliable modality input, while the mask and box prompts enable precise salient object localization.
Extensive experiments on three public datasets demonstrate that our method achieves state-of-the-art performance. Notably, HyPSAM has remarkable versatility, seamlessly integrating with different RGB-T SOD methods to achieve significant performance gains, thereby highlighting the potential of prompt engineering in this field. The code and results of our method are available at: https://github.com/milotic233/HyPSAM.

\end{abstract}

\begin{IEEEkeywords}
RGB-thermal, salient object detection, dynamic convolution, hybrid prompts, segment anything model.
\end{IEEEkeywords}

\begin{figure}[t]
\centering
\includegraphics[width=0.48\textwidth]{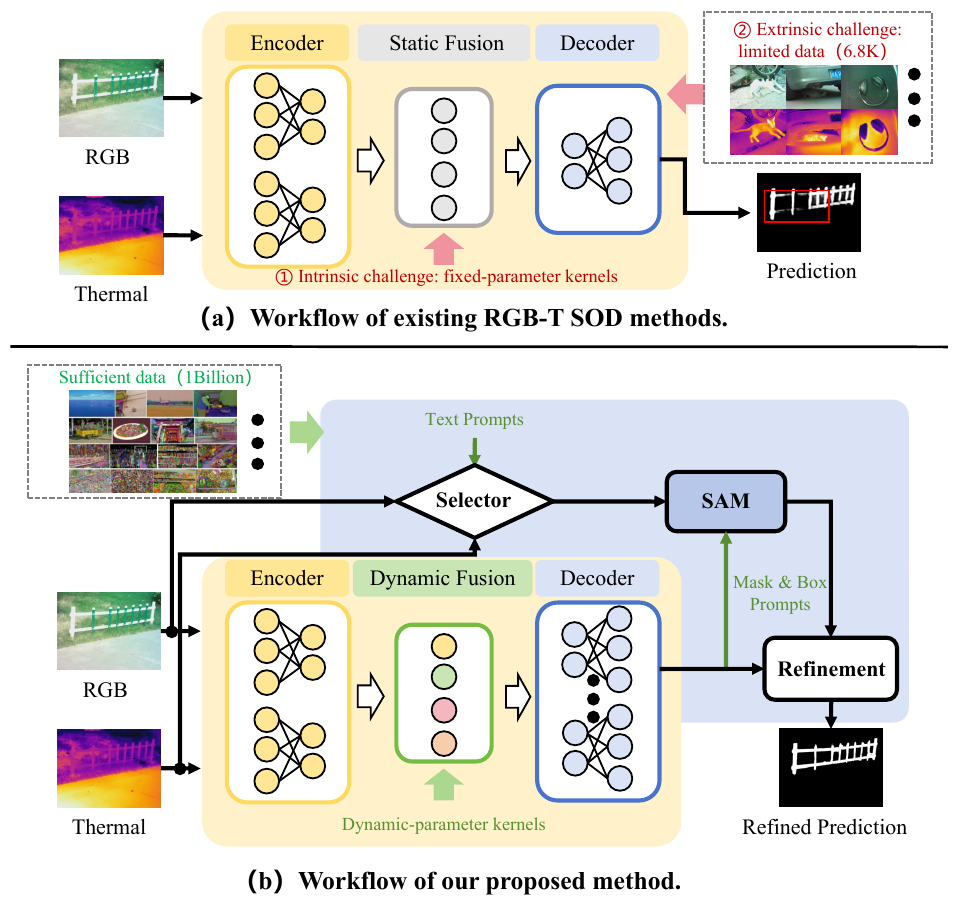}

\caption{Comparisons of the workflows of existing RGB-T SOD methods and our method. (a) Existing methods typically adopt the encoder-decoder paradigm, incorporating well-designed fusion strategies. (b) Our method designs a dynamic fusion network combined with hybrid prompts to guide SAM for accurate saliency predictions.}
\vspace{-0.5cm}
\label{fig:1}
\end{figure}

\section{Introduction}
\IEEEPARstart{S}{alient} object detection (SOD) aims to identify and segment the most visually attractive and prominent objects within a given scene~\cite{itti1998model}, which is foundational for numerous downstream applications, such as image retrieval~\cite{retrieval}, visual object tracking~\cite{mfg}, and camouflaged object detection~\cite{zhou2024decoupling}.
While unimodal SOD methods ~\cite{TGSOD, bbr,elsa, ELWNet} may suffer performance degradation in challenging scenarios, \emph{e.g.}, low illumination and cluttered backgrounds, recent studies have incorporated the complementary thermal infrared spectrum to enhance object perception, giving rise to the RGB-thermal (RGB-T) SOD task.

\begin{figure}[t]
    \centering
    \includegraphics[scale=0.55]{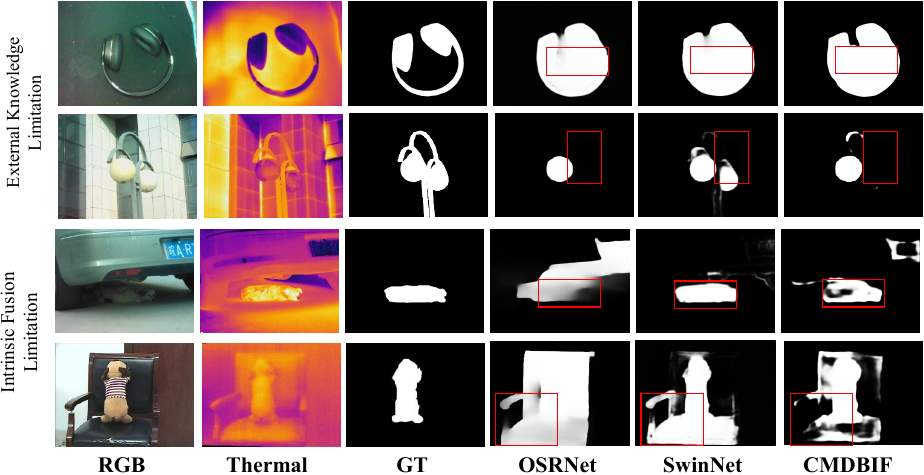}
    \caption{{Qualitative comparison illustrating two key limitations: external knowledge limitation (top two rows) leads to incomplete semantic understanding, while intrinsic fusion limitation (bottom two rows) causes inaccurate modality fusion. Red boxes highlight the incorrect saliency regions.}}
    \vspace{-0.5cm}
    \label{fig:2} 
\end{figure}

Existing works in RGB-T SOD, whether employing single stream~\cite{OSRNet}, dual stream~\cite{rgbd, wang2025progressive, Swinnet, MCFNet, CAVER, TNet, Uni, ASY, pos, cgfnet, revisit, MIDD, MENET, adnet, TIG, wav, uminet, mirror, 821, 1000, 5000, HDN, CSRNet, WGOFNet,wang2024learning,wang2024alignment,jin2025rethinking} and triple stream architectures~\cite {CMDBIF-Net}, share a common goal of exploring complementarity from dual modalities to improve the multi-modal representation. 
As illustrated in Fig.~\ref{fig:1}(a), most of them follow a similar workflow, but two key challenges remain unresolved.
First, the inherent differences between RGB and thermal images complicate the extraction of complementary features.  Existing fusion mechanisms often rely on static convolutional kernels and complex attention designs, which may lack adaptability to diverse scenes. 
Second, high-quality RGB-T annotations are costly to obtain, limiting available training data and making models prone to overfitting or poor generalization. As shown in Fig.~\ref{fig:2}, these issues frequently lead to incomplete object detection and blurred boundaries, especially in complex scenarios. 

In this paper, we propose a novel hybrid prompt-driven segment anything model (HyPSAM) that leverages the zero-shot generalization capabilities of the segment anything model (SAM)~\cite{SAM} for RGB-T SOD. As shown in Fig.\ref{fig:1}(b), HyPSAM integrates a coarse-to-fine framework by combining a robust saliency prediction network with a SAM-based refinement network.
However, directly applying SAM to RGB-T SOD is nontrivial due to two key limitations. First, as a general-purpose model, SAM relies on task-specific and well-designed prompts to accurately identify salient objects. Second, being trained solely on RGB images, SAM lacks inherent support for multi-modal inputs.

To this end, we introduce two core components. First, a dynamic fusion network (DFNet) is designed to improve intrinsic saliency detection by adaptively merging RGB and thermal features via context-aware dynamic convolutions and a multi-branch decoder. Second, a plug-and-play refinement network (P2RNet) is proposed to adapt SAM through hybrid prompts and modality-aware input selection. Specifically, a CLIP-based modality selector is used to choose the most informative modality based on scene semantics. Meanwhile, multi-level geometric prompts, derived from the initial saliency map, are introduced to guide SAM in refining predictions. 
Extensive experiments on three public RGB-T benchmarks demonstrate that HyPSAM outperforms state-of-the-art methods and its seamless integration with existing RGB-T SOD methods without task-specific training.

The main contributions are summarized as follows:

$\bullet$ We propose a novel hybrid prompt-driven framework that breaks the performance bottleneck by improving intrinsic feature fusion and embedding external rich semantic knowledge.

$\bullet$  We propose a dynamic fusion network that improves detection accuracy by employing context-aware dynamic convolutions and multi-branch decoding. It enables adaptive cross-modality interaction, overcoming the limitations of fixed-parameter kernels and improving feature fusion.

$\bullet$  We propose a plug-and-play refinement network to alleviate the limitations of data scarcity. By utilizing hybrid prompts, comprising text, masks, and boxes, it adapts SAM for RGB-T SOD tasks without additional training, achieving precise results and improved generalization.

\section{Related work}
\subsection{RGB-T Salient Object Detection}
RGB-T SOD aims to enhance scene understanding by incorporating the advantages of RGB and thermal modality, thus accurately segmenting the common saliency regions.
Recent advancements in this field have been largely driven by deep learning, with methods broadly categorized into three types based on architecture: single-stream, dual-stream, and triple-stream~\cite{OSRNet, CMDBIF-Net, Swinnet, MCFNet, CAVER, TNet, Uni, ASY, pos, cgfnet, revisit, MIDD, MENET, adnet, TIG, wav, uminet, mirror, 821, 1000, 5000, HDN, CSRNet, WGOFNet,wang2024learning,wang2024alignment,jin2025rethinking}. 

Among these, dual-stream architectures are the most prevalent due to their effectiveness in capturing cross-modal interactions. For example, Ma \emph{et al.}~\cite{MCFNet} proposed a modality complementary fusion network that mitigates the negative impact of low-quality modalities from both global and local perspectives. Liu \emph{et al.}~\cite{Swinnet} designed a dual-stream encoder based on the Transformer, which improves detection performance by optimizing cross-modal features through spatial and channel recalibration modules. 
More recently, Jin~\emph{et al.}~\cite{jin2025rethinking} proposed a lightweight local and global perception network that integrates convolutional inductive bias with Transformer-based global modeling to achieve efficient multimodal fusion.

Single-stream architectures are particularly advantageous for real-time applications. For example, OSRNet~\cite{OSRNet} aggregated multi-modal information at the early fusion stage using operations such as cascading and element-wise computation, achieving faster inference. 

Three-stream architectures, such as MDBIFNet~\cite{CMDBIF-Net}, utilize the third stream as an additional layer of interaction, enhancing the integration of complementary data from both modalities. The interaction branch strengthens cross-modal correlations, improving the overall performance of saliency detection.

Nevertheless, most existing RGB-T SOD methods rely heavily on static fusion modules with fixed parameters, which often struggle to adaptively capture the complex and dynamic modality relationships encountered in real-world scenarios. In contrast, our proposed framework introduces a dynamic fusion mechanism and incorporates hybrid prompt strategies to guide the foundation model, effectively alleviating modality bias and significantly enhancing generalization performance across diverse conditions.

\subsection{Dynamic Multi-Modal Fusion}
Dynamic fusion techniques have become increasingly important for multi-modal saliency detection due to their ability to adaptively model uncertain modality relationships. Unlike static fusion schemes with fixed parameters, dynamic fusion adjusts feature interactions based on input content, allowing more flexible and robust feature representation.

A major research direction is dynamic convolution, which generates input-dependent convolutional kernels to adaptively modulate feature extraction \cite{jia2016dynamic,wu2018dynamic,zhou2021decoupled}. 
For example, Pang \emph{et al.}\cite{HDF} proposed a hierarchical dynamic filtering network for cross-modal feature fusion, which significantly improves RGB-D salient object detection. Zhang \emph{et al.}~\cite{CCDF} developed an interlaced dynamic filtering network by decoupling dynamic convolution, dynamically combining discriminative RGB-D features to boost detection performance.
Beyond convolution, dynamic learning strategies have also been explored. For example, Jin \emph{et al.}~\cite{jin2025underwater} proposed a dual-stage self-paced learning framework combined with depth emphasis for underwater salient object detection.

Nonetheless, many dynamic filters remain RGB-centric and construct dynamic convolutional kernels along restricted dimensions, which limits their robustness. In contrast, our method adopts a simple yet effective dynamic convolution mechanism applied symmetrically to both modalities, which alleviates the issues caused by degraded quality.

\begin{figure*}[t]
    \centering
    \includegraphics[width=0.8\textwidth]{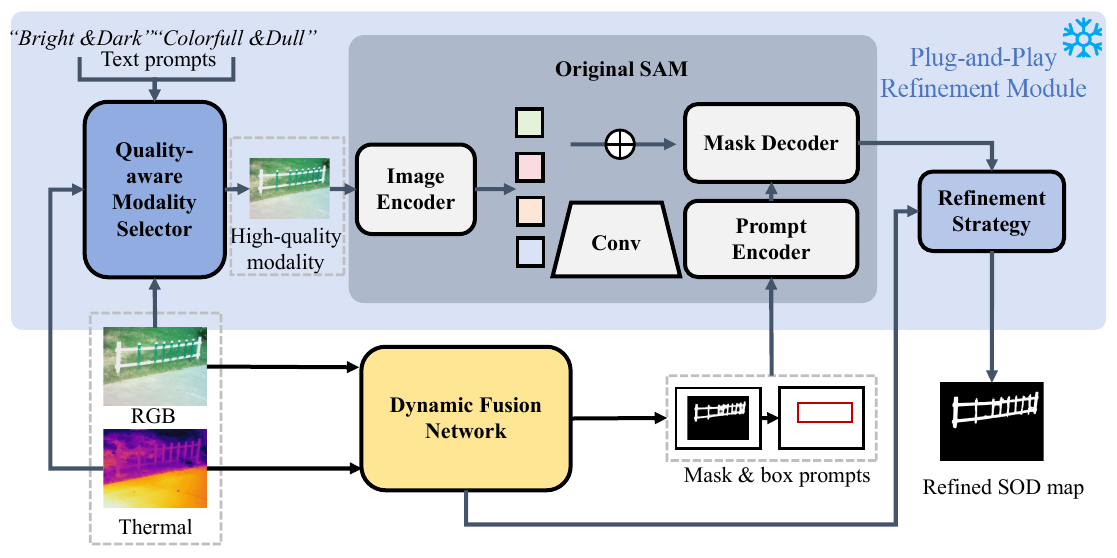}
    \caption{The framework of our proposed HyPSAM. We first feed the RGB-T image pairs into the DFNet to generate the saliency map, which is then used in conjunction with the augmentation strategy to create hybrid prompts. Additionally, the image pairs are processed by the quality-aware modality selector to select the most effective modality as input for SAM. Finally, the mask output from SAM and the saliency map are fused by the refinement strategy to obtain the final refined results.}
    \label{fig:3}
    \vspace{-0.5cm}
\end{figure*}  

\subsection{SAM and Its Application}
SAM~\cite{SAM}, trained on the SA-1B dataset with over one billion masks, has demonstrated exceptional visual representation. As a foundational model for image segmentation, it is renowned for its zero-shot transfer capabilities, allowing users to perform segmentation through simple prompts. 

Numerous studies have explored SAM's applications across various imaging domains. For example, 
Ma \emph{et al.}~\cite{ma2024segment} proposed MedSAM, which is designed for universal medical image segmentation, supported by a large-scale medical image dataset.
Wang \emph{et al.}~\cite{wang2024samrs} applied SAM to remote sensing, creating SAMRS, an architecture for generating large-scale remote sensing segmentation datasets.
Furthermore, SAM has also been extended to multi-modal vision tasks. For example,
Yu \emph{et al.}~\cite{yu2024exploring} proposed a depth-aware camouflage object detection and segmentation model that leverages the zero-shot capabilities of SAM to achieve precise segmentation in the RGB-D domain. 
Fang \emph{et al.}~\cite{fang2024semantic} developed a new RGB-T crowd-counting method using semantic maps from SAM to distinguish between foreground and background for guiding cross-modal feature fusion. Wu\emph{ et al.}~\cite{wu2025every} proposed SAGE, a multi-modality image fusion framework that distills semantic priors from SAM to balance visual quality and downstream task adaptability without relying on SAM during inference. Zhai \emph{et al.}~\cite{zhai2025weakly} propose a SAM-guided label optimization and progressive cross-modal cross-scale fusion framework for weakly supervised RGB-T salient object detection using scribble annotations.

While most existing works adopt adapter-based modifications or label generation, our HyPSAM directly utilizes SAM by integrating prompts with RGB-T saliency priors through a plug-and-play refinement framework without any additional task-specific training.

\section{Methodology}

The framework of HyPSAM is illustrated in Fig.~\ref{fig:3}, where DFNet and P2RNet are integrated to enhance generalization and improve saliency detection in complex scenarios. RGB-T image pairs are first processed by DFNet to generate hybrid prompts. Meanwhile, a quality-aware modality selector automatically determines the more reliable modality as the sole input to SAM, ensuring compatibility with its input format. Notably, SAM is used without any fine-tuning, with all components (image encoder, prompt encoder, and mask decoder) kept completely frozen. Hybrid prompts are injected exclusively through SAM’s standard prompt encoder, fully leveraging its inherent prompt-driven segmentation capability without modifying its architecture. Finally, a refinement strategy is applied to produce the final saliency map.

\begin{figure*}[t]
    \centering
    \includegraphics[width=0.9\textwidth]{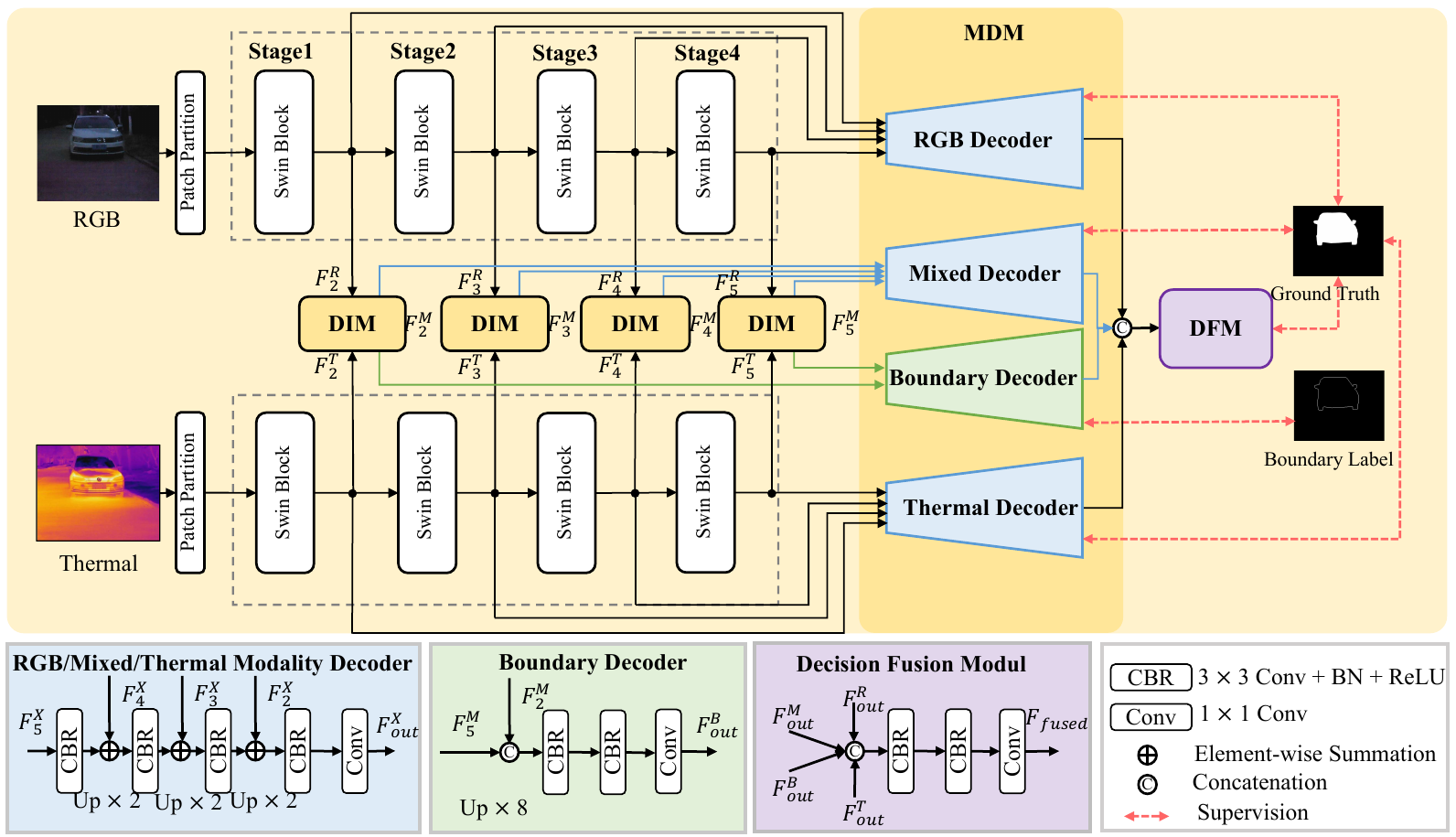}
    \caption{The architecture of DFNet. We feed the RGB-T image pairs into a Swin Transformer-based backbone to extract features. These features are then enhanced through the dynamic interaction module. Next, the multi-branch decoding module establishes dedicated branches for each modality to decode their respective features effectively. Finally, the decision fusion module combines the predictions from multiple branches, generating the saliency map.}
    \vspace{-0.5cm}
    \label{fig:4}
\end{figure*}

\begin{figure}[t]
    \centering
    \includegraphics[width=0.48\textwidth]{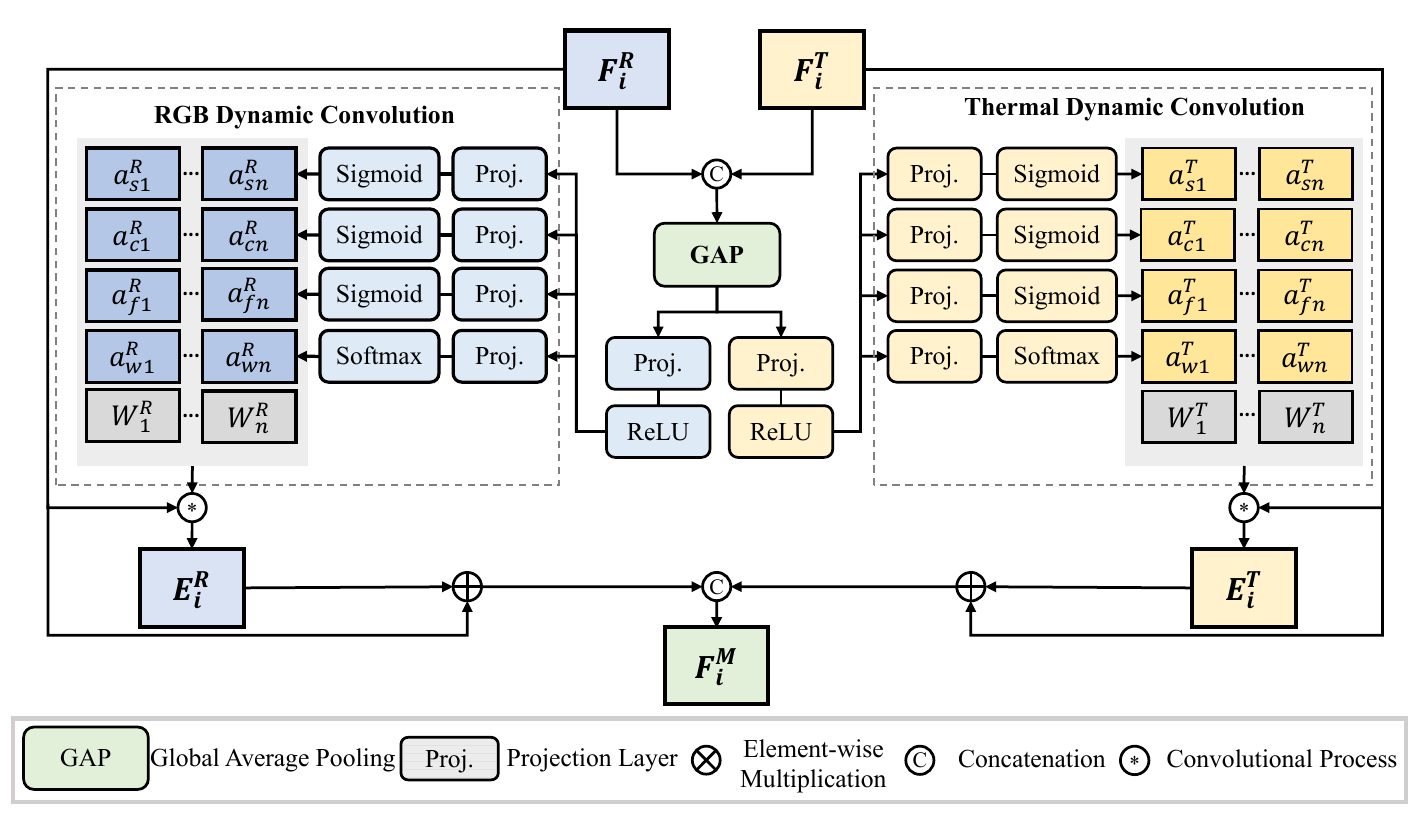}
    \caption{Details of the dynamic interaction module in DFNet. }
    \vspace{-0.5cm}
    \label{fig:5} 
\end{figure}  

\subsection{Dynamic Fusion Network}
Figure~\ref{fig:4} illustrates the architecture of DFNet.
We first utilize the Swin Transformer~\cite{swin} as the backbone to construct a symmetric dual-stream encoder, which combines the strengths of both Transformer and CNN, demonstrating robust capabilities in global and local feature representation. To reduce computational complexity, $1 \times 1$ convolutions compress the feature channels to 64, and only the last four layers of features are considered for interaction and decoding. Mathematically, the hierarchical RGB-T features are denoted as $ \{F_i^R|i=2,3,4,5\}$ and $ \{F_i^T|i=2,3,4,5\}$, respectively. 
Next, we design the dynamic interaction module to enhance multi-modal features $ \{F_i^M|i=2,3,4,5\}$ by dynamically integrating RGB and thermal cues, improving robustness in complex scenarios.
The multi-branch decoding module predicts single-modal and mixed-modal saliency results and boundary details. 
Finally, the decision fusion module merges the predictions from multiple branches and generates the initial saliency map.

{\emph{1) Dynamic Interaction Module:}}

To address the limitations of fixed convolutional kernels, we design a dynamic interaction module that adaptively enhances features for each modality. Unlike prior dynamic methods~\cite{CCDF, HDF} that typically produce filters conditioned on fused features to enhance the RGB stream alone, our method treats both RGB and thermal inputs equally, avoiding modality dominance bias.
Inspired by the omni-dimensional dynamic convolution (ODConv)~\cite{OMNI}, we employ multi-dimensional attention to generate dynamic convolutional weights tailored to each modality, as shown in Fig.~\ref{fig:5}.

Specifically, we first concatenate the modality-specific features to obtain the fused representation:
\begin{align}
F_i^f &= \text{Concat}(F_i^R, F_i^T). \label{eq:fuse}
\end{align}
where $F_i^R$ and $F_i^T$ denote the features from the RGB and thermal modalities, respectively, and $\text{Concat}(\cdot)$ indicates channel-wise concatenation.

Next, global contextual information is extracted by applying global average pooling (GAP) followed by a linear projection and non-linear activation:
\begin{align}
V_i^x &= \text{ReLU}(\text{Proj}^x(\text{GAP}(F_i^f))), \quad x \in \{R, T\}, \label{eq:2}
\end{align}
where $\text{GAP}(\cdot)$ denotes global average pooling, $\text{Proj}(\cdot)$ is linear projection layer, $\text{ReLU}(\cdot)$ is non-linear activation function.

The context vector $V_i$ is then used to compute dimension-specific attention scalars for each modality through multiple parallel projection branches.  For modality $x \in \{R, T\}$, the attention weights across four dimensions are calculated as:
\begin{equation}
\left\{
\begin{aligned}
a_{sn}^x &= \text{Sigmoid}(\text{Proj}_s^x(V_i^x)) \\
a_{cn}^x &= \text{Sigmoid}(\text{Proj}_c^x(V_i^x)) \\
a_{fn}^x &= \text{Sigmoid}(\text{Proj}_f^x(V_i^x)) \\
a_{wn}^x &= \text{Softmax}(\text{Proj}_w^x(V_i^x))
\end{aligned}
\right., \quad n = 1, 2, \dots, N
\end{equation}
where $N$ is the number of convolutional kernels, and Sigmoid or Softmax normalize the weights across corresponding dimensions, $a_{wn}^x$, $a_{fn}^x$, $a_{cn}^x$, and $a_{sn}^x$ correspond to kernel, filter, channel, and spatial attention weights, respectively.

The dimension-specific attention scalars are employed to adaptively reweight the convolutional parameters, enhancing cross-modal feature interactions in a fine-grained and effective manner. The context-aware dynamic convolution for each modality is then formulated as:
\begin{equation}
E_i^x = \sum_{n=1}^N \left( a_{wn}^x \odot a_{fn}^x \odot a_{cn}^x \odot a_{sn}^x \odot W_n^x \right) * F_i^x,
\end{equation}
where $F_i$ denotes input features, $E_i$ represents the enhanced features, $*$ indicates the convolution operation, while $\odot$ represents the multiplication along different dimensions. 
Here, $n$ denotes the index of the $n$-th convolutional kernel, with each kernel $W_n$ being modulated by the four attention dimensions before aggregation.

Finally, the enhanced mixed features are obtained by fusing the original input and enhanced features of both modalities:
\begin{equation}
F_i^M = \text{Concat}(E_i^R + F_i^R, \; E_i^T + F_i^T),
\end{equation}
where $E_i^R$ and $E_i^T$ indicate the enhanced features of RGB and thermal modalities, respectively.

{\emph{2) Multi-branch Decoding Module:}}
To address the challenges posed by modality quality discrepancies and false positive detection, the multi-branch decoding module enhances saliency prediction by decoupling the RGB and thermal modalities while preserving discriminative features. Unlike single-decoder methods, our module independently processes RGB, thermal, and mixed-modality features, enabling tailored predictions for each modality. By leveraging hierarchical fusion, our module progressively aggregates deep and shallow layer features, utilizing their complementary benefits to improve prediction accuracy.
The decoding process can be formulated as follows:

\begin{align}
{\tilde F_3^{x}}&=\text{Up}(\text{CBR}(\text{Up}(\text{CBR}(F_5^x))\oplus {F_4^x})) \oplus{F_3^x}, \label{eq:1}\\
{F_{out}^x}&=\text{CBR}(\text{Up}(\text{CBR}(\tilde{F_3^{x}}))\oplus{F_2^x}), x\in\{R,T,M\},\label{eq:2}
\end{align}
where ${F_{out}^x}$ and $\tilde{F_3^{x}}$ indicate the output and $3rd$ layer decoding features, $\text{CBR}$ means the sub-network consists of Convolution, Batch Normalization, and ReLU operations, $\text{Up}$ denotes bilinear interpolation, $\oplus$ is the element-wise addition, $x\in\{R,T,M\}$ represents the RGB, thermal and mixed-modality, respectively.

Moreover, to preserve the detailed edges in the saliency results, we develop a lightweight boundary decoder. Inspired by previous work~\cite{basnet, LDF}, we use the Canny operator for label decoupling, separating binary labels into boundary and content parts, and enabling supervised training on boundary details. Given that shallow features contain rich edge details but are also susceptible to noise, we fuse deep features from the $5th$ layer with shallow features from the $2nd$ layer to construct cross-level fused features for precise edge prediction. The boundary decoder is defined as follows:
\begin{equation}
{F_{out}^{B}}=\text{CBR}_{\times 2}(\text{Concat}(\text{Up}(F_5^M), F_2^M))),
\end{equation}
where ${F_{out}^{B}}$ is the boundary decoding features, $\text{CBR}_{\times 2}$ indicates the $\text{CBR}$ is stacked two times.

{\emph{3) Decision Fusion Module:}}
In addition, we introduce a decision fusion module to produce a final saliency map that aggregates the decoding features from multiple branches. This module is jointly trained with the decoders described above. The details of the architecture are illustrated in Fig.~\ref{fig:3}. We first concatenate the output features from the four branches. Then, we apply convolutional layers to fuse different features. The fused saliency result can be described as follows:
\begin{equation}
{F_{f}}=\text{CBR}_{\times 2}(\text{Concat}(F_{out}^M, F_{out}^R,F_{out}^T,F_{out}^B)).
\end{equation}

{\emph{4) Loss Function:}}
To strengthen the ability of the multi-branch decoding module to capture discriminative features, different loss functions are assigned to each branch for supervision.
For the boundary prediction, class imbalance is a critical challenge due to the sparse nature of edge pixels in the overall image. To address this, the dice loss~\cite{dice} is used to supervise the boundary decoder. Additionally, the other decoders are supervised by the hybrid loss~\cite{basnet}:
\begin{equation}
{\ell}_{hyb} = {\ell}_{bce}+ {\ell}_{ssim}+{\ell}_{iou},		
\end{equation}
where ${\ell}_{bce}$, ${\ell}_{ssim}$ and ${\ell}_{iou}$ denote BCE loss~\cite{bce}, SSIM loss~\cite{ssim} and IoU loss~\cite{iou}, respectively. The total loss $\mathcal{L}$ is expressed as follows:
\begin{equation}
\mathcal{L} = {\ell}_{hyb}^{R}+ {\ell}_{hyb}^{T}+{\ell}_{hyb}^{M}+{\ell}_{dice}^{B}+{\ell}_{hyb}^{F},		
\end{equation}
where ${\ell}_{hyb}^{R}$, ${\ell}_{hyb}^{T}$, ${\ell}_{hyb}^{M}$, ${\ell}_{dice}^{B}$ and ${\ell}_{hyb}^{F}$ indicate RGB, thermal, mixed-modality, boundary and fusion loss, respectively.

\subsection{Plug-and-Play Refinement Network}
Although the proposed DFNet has achieved excellent performance on publicly available RGB-T datasets, its generalizability is restricted by the limited scale and diversity of the training data. Consequently, it still encounters misclassification in challenging or unseen scenarios. Such inaccuracies in saliency maps hinder the practical application of SOD methods, as precise and reliable outputs are essential for real-world interaction. 

Existing approaches~\cite{gao2024multi, 10443051} often incorporate the SAM encoder as a backbone or design adapters to enhance feature extraction. Nevertheless, these manners usually incur additional training costs. To address this problem, we propose a P2RNet, which serves as a general optimization network to further refine coarse saliency maps, as shown in Fig.~\ref{fig:3} (blue background). P2RNet aims to simply leverage SAM from multiple perspectives without further modifying its weights or architecture. 
The refinement process begins with RGB-T image pairs being processed by a quality-aware modality selector, which evaluates the quality of both modalities and selects the optimal one for further processing. 
Based on this evaluation, the augmentation
strategy generates high-quality visual prompts, including masks and bounding boxes, derived from the initial saliency map. These hybrid prompts guide the SAM to precisely distinguish the saliency object. The final saliency maps are refined by incorporating both the initial predictions and segmentation masks, resulting in sharper boundaries and more complete object structures.

\begin{figure}[t]
    \centering
    \includegraphics[scale=0.38]{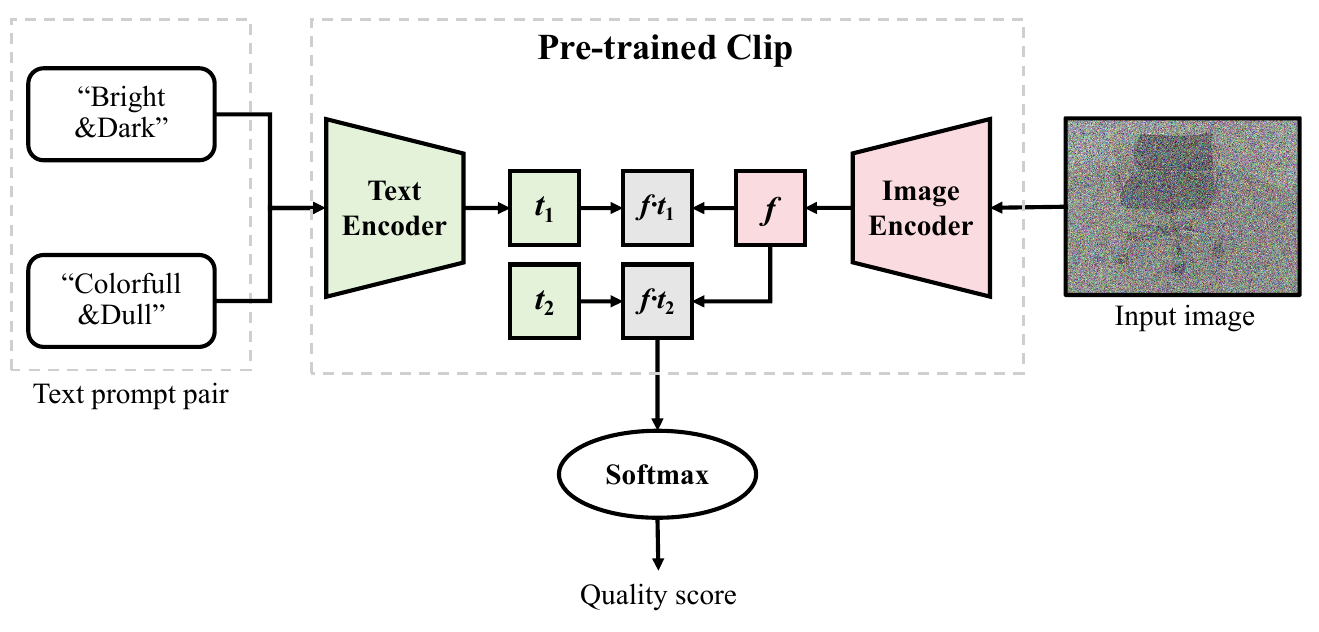}
    \caption{Details of the quality-aware modality selector. }
    \vspace{-0.2cm}
    \label{fig:6} 
\end{figure}

{\emph{1) Quality-aware Modality Selector:}} Since RGB-T images are well-aligned and represent the same scene, both modalities share inherent semantic similarities, allowing SAM to process thermal images to some extent. Therefore, we attempt to feed the reliable quality modality into SAM and achieve good segmentation results without any task-specific fine-tuning. 
Nevertheless, image degradation caused by various factors makes accurate quality assessment challenging, thus, a simple binary classifier explicitly trained on limited labeled datasets may fail to reliably evaluate image quality (IQA). Inspired by advances in image quality assessment, we directly employ CLIP-IQA~\cite{wang2023exploring} as a quality-aware modality selector, as shown in Fig.~\ref{fig:6}. CLIP is a foundational model trained in contrastive learning, primarily used for the cross-modal alignment of language and images. CLIP-IQA applies the large model to the IQA task, considering both quality and abstract perception. It calculates the cosine similarity between the image features and antonym prompts and then uses Softmax to compute the final quality score. 
\begin{equation}
{s_i}=\frac{f \odot t_i} {{\Vert f \Vert} \cdot {\Vert t_i \Vert}}, i\in\{1,2\} ,
\end{equation}
where $f$ is the image features and $t_i$ are the features from antonym prompts, $\odot$ is the dot product and $\Vert \cdot \Vert$ denotes ${\ell}_{2}$ norm.

\begin{equation}
{s}=\frac{e^{s_1}} {e^{s_1} + e^{s_2}} ,
\end{equation}
where $s_1$ and $s_2$ are the scores corresponding to the positive and negative quality descriptions, respectively. The exponentiation operation amplifies the difference between the two similarity scores, making the output score more sensitive to the dominant attribute.

To enhance reliability, we adopt two pairs of antonym prompts, \emph{i.e.}, “bright and dark” and “colorful and dull”, to reduce the ambiguity of the prompt. The “bright and dark” prompt assesses the illumination conditions of the scene. If the confidence score $s_{\alpha}$ exceeds the threshold $\tau$, the RGB image is determined to be more reliable. The “colorful and dull” prompt evaluates the richness of the scene information. If the quality score $s_{\beta}$ exceeds the threshold $\theta$, the RGB image is of higher quality. Otherwise, the thermal modality is selected. The effective combination of two text prompts allows for a robust selection of the most suitable modality.

\begin{equation}
m_{re} = 
\begin{cases} 
m_r & \text{if } s_{\alpha} > \tau \text{ or } s_{\beta} > \theta, \\
m_t & \text{others. } 

\end{cases}
\end{equation}

\begin{figure}[t]
    \centering
    \includegraphics[scale=0.58]{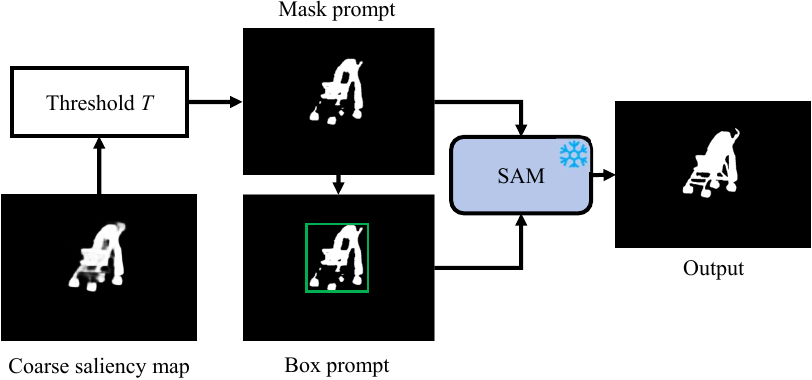}
    \caption{Illustration of saliency prompts generation.}
    \vspace{-0.5cm}
    \label{fig:7} 
\end{figure}  

{\emph{2) Prompt Augmentation Strategy:}} SAM is an interactive segmentation model that supports diverse types of prompts, including points, boxes, text, and masks. Previous work, such as SAMAug~\cite{dai2023samaug}, has demonstrated that augmenting point-based prompts can boost segmentation performance, underscoring the potential of visual prompt engineering. However, relying exclusively on sparse point prompts may result in incomplete objects, particularly for complex or irregular object shapes. 

To address this limitation, we propose a hybrid prompt augmentation strategy that combines both mask and box prompts, providing comprehensive global and local guidance for salient object representation.
Specifically, we begin with a coarse saliency map, which serves as an initial approximation of object regions. Given that this map may contain noise and exhibit ambiguous boundaries, especially near object edges, we apply a binarization operation with a threshold $T$ to produce a refined binary mask $S_t$, which helps suppress uncertain regions and reduces the impact of blurred contours on SAM’s segmentation results.
We then use $S_t$ as a mask prompt by integrating it into the image embedding via element-wise addition, allowing SAM to attend to spatially informative regions. Furthermore, we extract bounding boxes around the $S_t$ to generate box prompts. These box-level cues impose additional spatial constraints, which enhance boundary precision and improve segmentation robustness. The entire hybrid prompt generation process is illustrated in Fig.~\ref{fig:7}.

{\emph{3) Refinement Strategy:}} While hybrid prompts improve SAM's segmentation accuracy, complex saliency regions can still lead to ambiguities, such as hollow objects. To address this issue, we propose a refinement strategy designed to preserve both the salient regions while sharpening edges.
Therefore, we merge the coarse saliency map $S_t$ with the segmentation mask $S_g$ to produce the refined map ${S_r}$, maintaining the integrity of the objects. The refinement process is described as follows:
\begin{equation}
{S_r(x,y)}=\max\{{S_t(x,y)},{S_g(x,y)}\},
\end{equation}
where the maximum operation is applied element-wise across the two maps, combining their strengths for optimal saliency representation.

\section{Experiments}
\subsection{Datasets and Evaluation Metrics}
We train and test our method on three well-known RGB-T benchmark datasets, namely VT821~\cite{821}, VT1000~\cite{1000}, and VT5000~\cite{5000}. 
Specifically, the VT821 dataset includes 821 pairs of well-aligned RGB-T images and their corresponding ground-truth labels, covering approximately 60 scenes, and some samples have added noise interference to enhance the challenge of the dataset.
The VT1000 dataset expands the VT821 dataset, containing 1,000 pairs of RGB-T images and ground-truth labels collected in real-life scenarios.
The VT5000 dataset consists of 5,000 pairs of RGB-T images, with significantly enhanced diversity in objects and scenes, which includes 13 challenging attributes, \emph{i.e.}, big salient object (BSO), center bias (CB), cross image boundary (CIB), image clutter (IC), low illumination (LI), multiple salient objects (MSO), out of focus (OF), small salient object (SSO), similar appearance (SA), thermal cross (TC), bad weather (BW), bad RGB (bRGB), and bad thermal (bT), respectively.

Following the common setting in these methods ~\cite{Swinnet,adnet}, we select 2,500 image pairs from VT5000 as the training set, while the remaining 2,500 image pairs in VT5000, along with the VT821 and VT1000 datasets, constitute the testing set.
We evaluate the performance of the model using widely adopted metrics~\cite{eva, eva1, eva2, eva3,eva4}, including mean F-measure $(F_{avg})$, maximum F-measure $(F_{max})$, weighted F-measure $(F_w)$, MAE ($\mathcal{M}$), E-measure $(E_m)$, S-measure $(S_m)$, and Precision-Recall curve.

The F-measure is a comprehensive metric that evaluates the balance between precision and recall, defined as:
\begin{equation}
F_{m}=(\beta^{2}+1)\cdot{\frac{precision\cdot recall}{\beta^{2}\cdot precision+recall}},
\end{equation}
where $\beta^{2}$ is empirically set to 0.3 to emphasize the precision over recall. 

The $\mathcal{M}$ quantifies the average absolute error between the predicted map and ground truth, defined as:
\begin{equation}
\mathcal{M}={\frac{1}{W\times H}}\sum_{x=1}^{W}\sum_{y=1}^{H}|S(x,y)-G(x,y)|,
\end{equation}
where $S$ and $G$ denote the predicted saliency map and GT, respectively. $W$ and $H$ represent the width and height, respectively. $(x,y)$ denotes the pixel coordinates.

$E_m$ evaluates both global and local similarity between the predicted saliency map and the ground truth, defined as:
\begin{equation}
E_{m}=\frac{1}{W\times H}\sum_{x=1}^{W}\sum_{y=1}^{H}\varphi\bigl({S}(x,y),{G}(x,y)\bigr),
\end{equation}
where $\varphi$ denotes the matrix entry.

$S_{m}$ evaluates the structural similarity between the ground truth and the predicted saliency map, defined as:
\begin{equation}
S_{m}=\alpha S_{o}+\left(1-\alpha\right)S_{r},
\end{equation}
where $S_{o}$ and $S_{r}$ represent the object-aware structural similarity and the region-aware structural similarity, respectively. The weight $\alpha$ is set to 0.5 to balance these two components.

\subsection{Implementation Details}
HyPSAM is implemented on the PyTorch platform with an Intel i9 13900K CPU and dual NVIDIA RTX 4090 GPUs. The input samples are resized to a resolution of $384\times384$. During the training phase, the training samples are augmented in various ways, including random flipping, rotating, and clipping. The backbone of DFNet is initialized with a pre-trained SwinV2-B network~\cite{swin}. Additionally, we train the network for 50 epochs with a batch size of 8 using the SGD optimizer. The learning rate for the backbone is set to $5e^{-3}$, while the learning rate of other parameters is set to $5e^{-2}$ and scheduled by the cosine strategy. 
The thresholds $\tau$ and $\theta$ in the quality-aware selector are set to 0.01 and 0.85, respectively. The remaining parameter settings follow CLIP-IQA. We select ViT-H as the pre-trained weights for SAM.

\begin{table*}[htbp]
\setlength\tabcolsep{2pt}
\scriptsize
\centering
\caption{Quantitative comparison with different RGB-T methods. The best and second-best results are highlighted in \textbf{bold} and \underline{underline}. $\uparrow$ ($\downarrow$) denotes larger is better (smaller is better). `$-$' indicates the code or result is not available.}
\resizebox{\textwidth}{!}{
\begin{tabular}{r|c|cccccc|cccccc|cccccc}
\toprule
\multirow{2}[4]{*}{Methods} & \multirow{2}[4]{*}{Pub.Info} & \multicolumn6{c|}{VT5000} & \multicolumn6{c|}{VT1000} & \multicolumn{6}{c}{VT821} \\
& & $F_{{avg}}\uparrow$ & $F_{{max}}\uparrow$ & $F_w\uparrow$ & $\mathcal{M} \downarrow$ & $E_m\uparrow$ & $S_m\uparrow$ & $F_{{avg}}\uparrow$ & $F_{{max}}\uparrow$ & $F_w\uparrow$ & $\mathcal{M} \downarrow$ & $E_m\uparrow$ & $S_m\uparrow$ & $F_{{avg}}\uparrow$ & $F_{{max}}\uparrow$ & $F_w\uparrow$ & $\mathcal{M} \downarrow$ & $E_m\uparrow$ & $S_m\uparrow$ \\
\midrule
SGDLL~\cite{1000} & $\text{TMM20}$ & 0.672 & 0.737 & 0.558 & 0.089 & 0.824 & 0.750 & 0.764 & 0.807 & 0.652 & 0.090 & 0.856 & 0.787 & 0.731 & 0.780 & 0.583 & 0.085 & 0.846 & 0.764 \\
CSRNet~\cite{CSRNet} & $\text{TCSVT21}$ & 0.811 & 0.857 & 0.796 & 0.042 & 0.905 & 0.868 & 0.877 & 0.918 & 0.878 & 0.024 & 0.925 & 0.918 & 0.831 & 0.880 & 0.821 & 0.038 & 0.909 & 0.885 \\
CGFNet~\cite{cgfnet} & $\text{TCSVT22}$ & 0.851 & 0.887 & 0.831 & 0.035 & 0.922 & 0.883 & 0.906 & 0.936 & 0.900 & 0.023 & 0.944 & 0.923 & 0.845 & 0.885 & 0.829 & 0.038 & 0.912 & 0.881 \\
SwinNet~\cite{Swinnet} & $\text{TCSVT22}$ & 0.865 & 0.915 & 0.846 & 0.026 & 0.942 & 0.912 & 0.896 & 0.948 & 0.894 & 0.018 & 0.947 & 0.938 & 0.847 & 0.903 & 0.818 & 0.030 & 0.926 & 0.904 \\
ADFF~\cite{5000} & $\text{TMM22}$ & 0.778 & 0.863 & 0.722 & 0.048 & 0.891 & 0.864 & 0.847 & 0.923 & 0.804 & 0.034 & 0.921 & 0.91 & 0.717 & 0.804 & 0.627 & 0.077 & 0.843 & 0.810 \\
TNet~\cite{TNet} & $\text{TMM22}$ & 0.846 & 0.895 & 0.840 & 0.033 & 0.927 & 0.895 & 0.889 & 0.937 & 0.895 & 0.021 & 0.937 & 0.929 & 0.842 & 0.904 & 0.841 & 0.030 & 0.919 & 0.899 \\
OSRNet~\cite{OSRNet} & $\text{TIM22}$ & 0.823 & 0.866 & 0.807 & 0.040 & 0.908 & 0.875 & 0.892 & 0.929 & 0.891 & 0.022 & 0.935 & 0.926 & 0.814 & 0.862 & 0.801 & 0.043 & 0.896 & 0.875 \\
ACMANet~\cite{ASY} & $\text{KBS22}$ & 0.858 & 0.890 & 0.823 & 0.033 & 0.932 & 0.887 & 0.904 & 0.933 & 0.889 & 0.021 & 0.945 & 0.927 & 0.837 & 0.873 & 0.807 & 0.035 & 0.914 & 0.883 \\
MCFNet~\cite{MCFNet} & $\text{AI22}$ & 0.848 & 0.886 & 0.836 & 0.033 & 0.924 & 0.887 & 0.902 & 0.939 & 0.906 & 0.019 & 0.944 & 0.932 & 0.844 & 0.889 & 0.835 & 0.029 & 0.918 & 0.891 \\
CMDBIF~\cite{CMDBIF-Net} & $\text{TCSVT23}$ & 0.868 & 0.892 & 0.846 & 0.032 & 0.933 & 0.886 & 0.914 & 0.931 & 0.909 & 0.019 & 0.952 & 0.927 & 0.856 & 0.887 & 0.837 & 0.032 & 0.923 & 0.882 \\
CAVER~\cite{CAVER} & $\text{TIP23}$ & 0.856 & 0.897 & 0.849 & 0.028 & 0.935 & 0.899 & 0.906 & 0.945 & 0.912 & 0.016 & 0.949 & 0.938 & 0.854 & 0.897 & 0.846 & 0.026 & 0.928 & 0.897 \\
ADNet~\cite{adnet} & $\text{MMA23}$ & 0.893 & 0.924 & 0.884 & 0.022 & 0.953 & 0.924 & 0.916 & 0.952 & 0.920 & 0.015 & 0.952 & 0.944 & 0.869 & 0.915 & 0.860 & 0.024 & 0.930 & 0.915 \\
WGOFNet~\cite{WGOFNet} & $\text{TOMM24}$ & 0.883 & 0.912 & 0.873 & 0.025 & 0.945 & 0.911 & 0.919 & 0.946 & 0.922 & 0.016 & 0.951 & 0.940 & 0.875 & 0.911 & 0.868 & 0.025 & 0.934 & 0.908 \\
UMINet~\cite{uminet} & $\text{TVC24}$ & 0.831 & 0.877 & 0.820 & 0.035 & 0.919 & 0.882 & 0.892 & 0.935 & 0.896 & 0.021 & 0.941 & 0.926 & 0.791 & 0.849 & 0.782 & 0.054 & 0.879 & 0.905 \\
TCINet~\cite{lv2024transformer} & $\text{TCE24}$ & {0.905} & 0.927 & {0.900} & \textbf{0.019} & \underline{0.959} & 0.925 & {0.925} & 0.943 & 0.928 & 0.014 & 0.956 & 0.944 & 0.882 & 0.910 & 0.879 & \underline{0.021} & 0.942 & 0.915 \\
LAFB~\cite{wang2024learning} & $\text{TCSVT24}$ & 0.857 & 0.893 & 0.841 & 0.030 & 0.931 & 0.893 & 0.905 & 0.937 & 0.905 & 0.018 & 0.945 & 0.932 & 0.843 & 0.884 & 0.817 & 0.034 & 0.915 & 0.884 \\
SACNet~\cite{wang2024alignment} & $\text{TMM24}$ & 0.901 & 0.922 & 0.888 & 0.021 & 0.957 & 0.917 & 0.923 & 0.949 & 0.927 & 0.014 & \underline{0.958} & 0.942 & 0.868 & 0.904 & 0.859 & 0.025 & 0.932 & 0.906 \\
DSCDNet~\cite{yu2024dual} & $\text{TCE24}$ & 0.888 & $\text{-}$ & 0.881 & 0.023 & 0.949 & 0.918 & 0.921 & $\text{-}$ & 0.927 & 0.014 & 0.955 & 0.946 & 0.876 & $\text{-}$ & 0.873 & 0.022 & 0.940 & 0.915 \\
FFANet~\cite{zhou2024frequency} & $\text{PR24}$ & 0.886 & $\text{-}$ & $\text{-}$ & 0.021 & 0.953 & 0.918 & 0.918 & $\text{-}$ & $\text{-}$ & 0.014 & 0.955 & 0.943 & 0.855 & $\text{-}$ & $\text{-}$ & 0.027 & 0.926 & 0.905 \\
PATNet~\cite{jiang2024patnet} & $\text{KBS24}$ & 0.883 & 0.916 & 0.879 & 0.023 & 0.946 & 0.917 & 0.910 & 0.948 & 0.920 & 0.015 & 0.951 & 0.940 & 0.870 & 0.914 & 0.872 & 0.024 & 0.933 & 0.910 \\
ISMNet~\cite{wang2025intra} & $\text{TCSVT25}$ & 0.885 & $\text{-}$ & 0.876 & 0.025 & 0.945 & 0.913 & 0.922 & $\text{-}$ & 0.924 & 0.014 & 0.954 & 0.942 & \underline{0.886} & $\text{-}$ & \underline{0.881} & \underline{0.021} & \underline{0.945} & 0.917 \\
ConTriNet~\cite{tang2025divide} & $\text{TPAMI25}$ & 0.898 & 0.927 & 0.895 & \underline{0.020} & 0.956 & 0.923 & 0.917 & 0.943 & 0.923 & 0.015 & 0.953 & 0.941 & 0.878 & 0.914 & 0.875 & 0.022 & 0.940 & 0.916 \\
KAN-SAM~\cite{li2025kan}  & $\text{ICME25}$ &    \underline{0.909} &{0.931} & \underline{0.905} & \underline{0.020} & {0.957} &{0.927} & \underline{0.930} & {0.947} & \underline{0.934} & \underline{0.013} & \underline{0.958} & {0.946} & {0.883} & {0.911} & {0.880} &{0.025} & {0.932} &{0.915}\\

 \midrule
DFNet & $\text{-}$ & 0.899 & \underline{0.933} & 0.898 & \underline{0.020} & 0.958 & \underline{0.930} & 0.920 & \underline{0.956} & {0.930} & \underline{0.013} & 0.955 & \underline{0.950} & 0.879 & \underline{0.925} & \underline{0.881} & 0.022 & 0.940 & \underline{0.926} \\
HyPSAM & $\text{-}$ & \textbf{0.928} & \textbf{0.939} & \textbf{0.911} & \textbf{0.019} & \textbf{0.963} & \textbf{0.931} & \textbf{0.946} & \textbf{0.957} & \textbf{0.944} & \textbf{0.011} & \textbf{0.965} & \textbf{0.954} & \textbf{0.914} & \textbf{0.930} & \textbf{0.903} & \textbf{0.020} & \textbf{0.948} & \textbf{0.932} \\
\bottomrule
\end{tabular}}%
\label{tab:1}%
\end{table*}

\subsection{Comparison with the State-of-the-art Methods}
To evaluate the performance of the proposed HyPSAM, we compare it with 23 state-of-the-art RGB-T methods on public benchmarks, namely, SGDL~\cite{1000}, CSRNet~\cite{CSRNet}, CGFNet~\cite{cgfnet}, SwinNet~\cite{Swinnet}, ADF~\cite{5000}, TNet~\cite{TNet}, OSRNet~\cite{OSRNet}, ACMANet~\cite{ASY}, MCFNet~\cite{MCFNet}, CMDBIF~\cite{CMDBIF-Net}, CAVER~\cite{CAVER}, ADNet~\cite{adnet}, WGOFNet~\cite{WGOFNet}, UMINet~\cite{uminet}, TCINet~\cite{lv2024transformer}, LAFB~\cite{wang2024learning}, SACNet~\cite{wang2024alignment}, DSCDNet~\cite{yu2024dual}, FFANet~\cite{zhou2024frequency}, PATNet~\cite{jiang2024patnet}, ISMNet~\cite{wang2025intra}, ConTriNet~\cite{tang2025divide} and KAN-SAM~\cite{li2025kan}. For a fair comparison, the saliency maps used for the comparison are provided by published works, and the metrics are calculated using the same evaluation toolbox~\cite{mirror}.

\emph{1) Quantitative Comparison:}
Table~\ref{tab:1} presents the comprehensive quantitative comparison of our method against the state-of-the-art approaches on RGB-T benchmarks.
The results demonstrate that HyPSAM achieves significant performance improvements across all benchmarks, surpassing existing methods by a substantial margin.

\begin{figure*}[t]
    \centering

    {\includegraphics[scale=0.38]{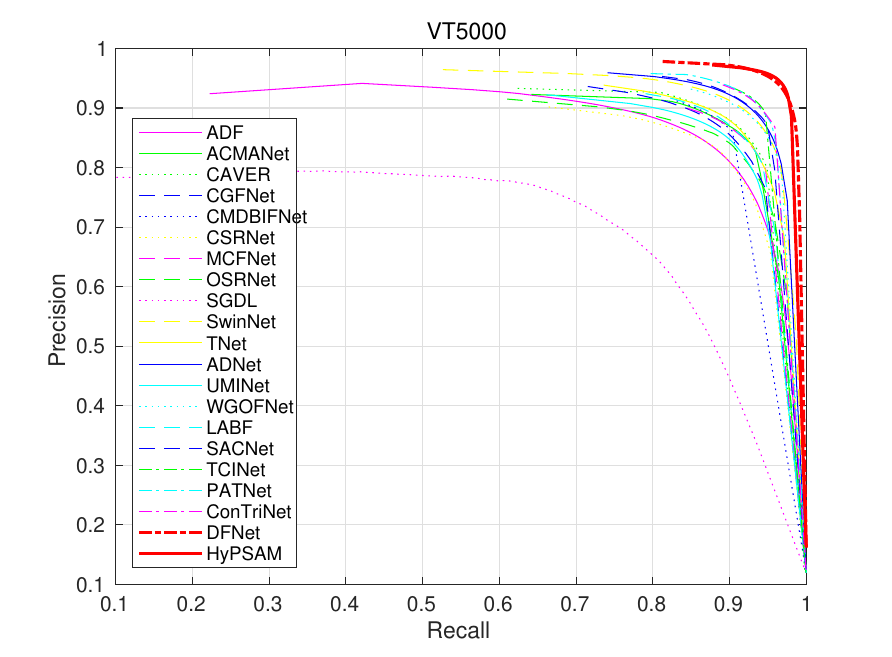}
    \includegraphics[scale=0.38]{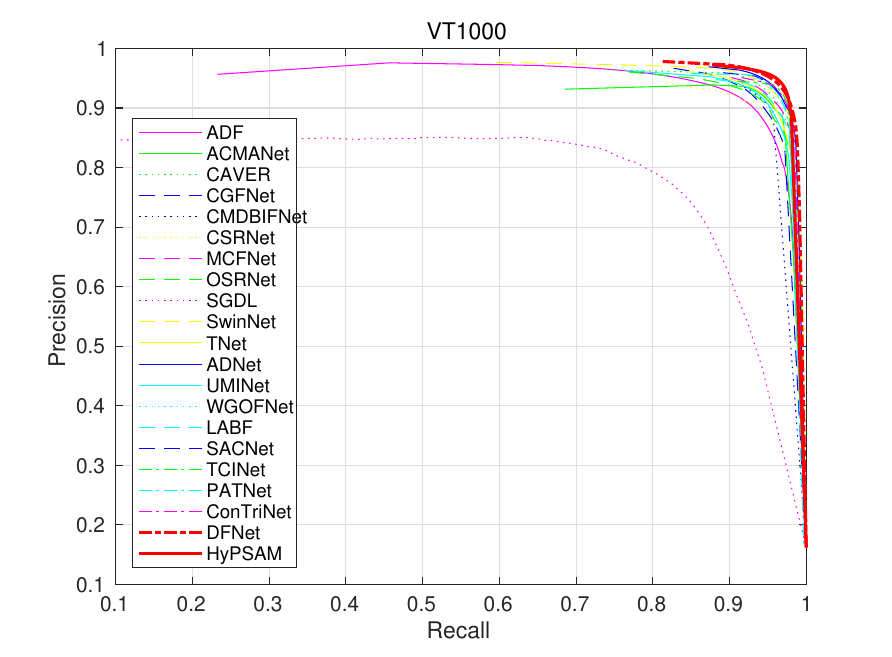}
    \includegraphics[scale=0.38]{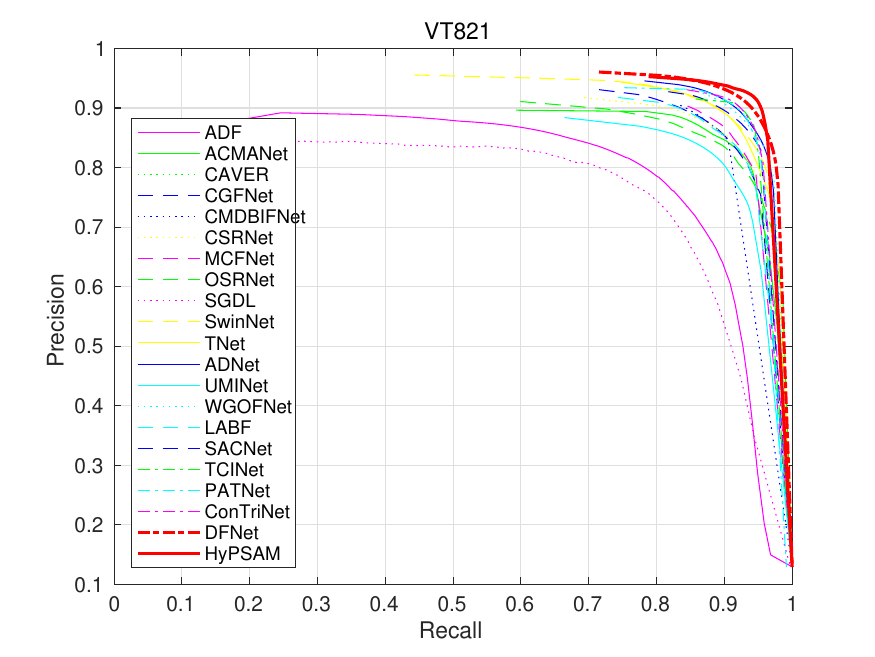}
    }

    \caption{Precision (vertical axis) - Recall (horizontal axis) curves comparison with the state-of-the-art RGB-T SOD methods on VT5000, VT1000, and VT821 datasets. The red solid curves show that our method outperforms the existing models.}
    \vspace{-0.3cm}
    \label{fig:8}
\end{figure*}

To provide a deeper analysis, we compare our HyPSAM with representative encoder-decoder-based methods. Specifically, compared to OSRNet~\cite{OSRNet}, a representative single-stream method, the $F_w$ metric, a critical measure for evaluating segmentation accuracy, shows significant improvements with HyPSAM, increasing by 10.4\%, 5.3\%, and 10.2\% on the VT5000, VT1000, and VT821 datasets, respectively. These experimental results demonstrate that our method effectively integrates complementary information rather than simplistic single-stream fusion.
In comparison to SwinNet~\cite{Swinnet}, a dual-stream architecture optimized for edge details, HyPSAM achieves notable improvements in $F_w$, $E_m$, and $S_m$ on the VT5000 dataset by 6.5\%, 2.1\%, and 1.9\%, respectively. These results demonstrate the superiority of our approach in enhancing dynamic feature interaction and preserving object boundary details. 
Compared to the triple-stream CMDBIF~\cite{CMDBIF-Net}, HyPSAM improved the $S_m$ by 4.5\%, 2.7\%, and 5.0\% on three benchmarks, respectively. This indicates that the saliency objects predicted by HyPSAM exhibit superior structural integrity.  
Moreover, compared with recent advanced methods \emph{e.g.}, {ISMNet}~\cite{wang2025intra}, and ConTriNet~\cite{tang2025divide}, our method adaptively fuses complementary cues and absorbs semantic features provided by SAM, yielding satisfactory detection results.
In Fig.~\ref{fig:8}, we display the Precision-Recall curves on three RGB-T SOD datasets, which show that our method can obtain the highest precision on the three datasets.

\begin{figure*}[t]
\setlength\tabcolsep{2.5pt}
    \centering
    \includegraphics[width=\linewidth]{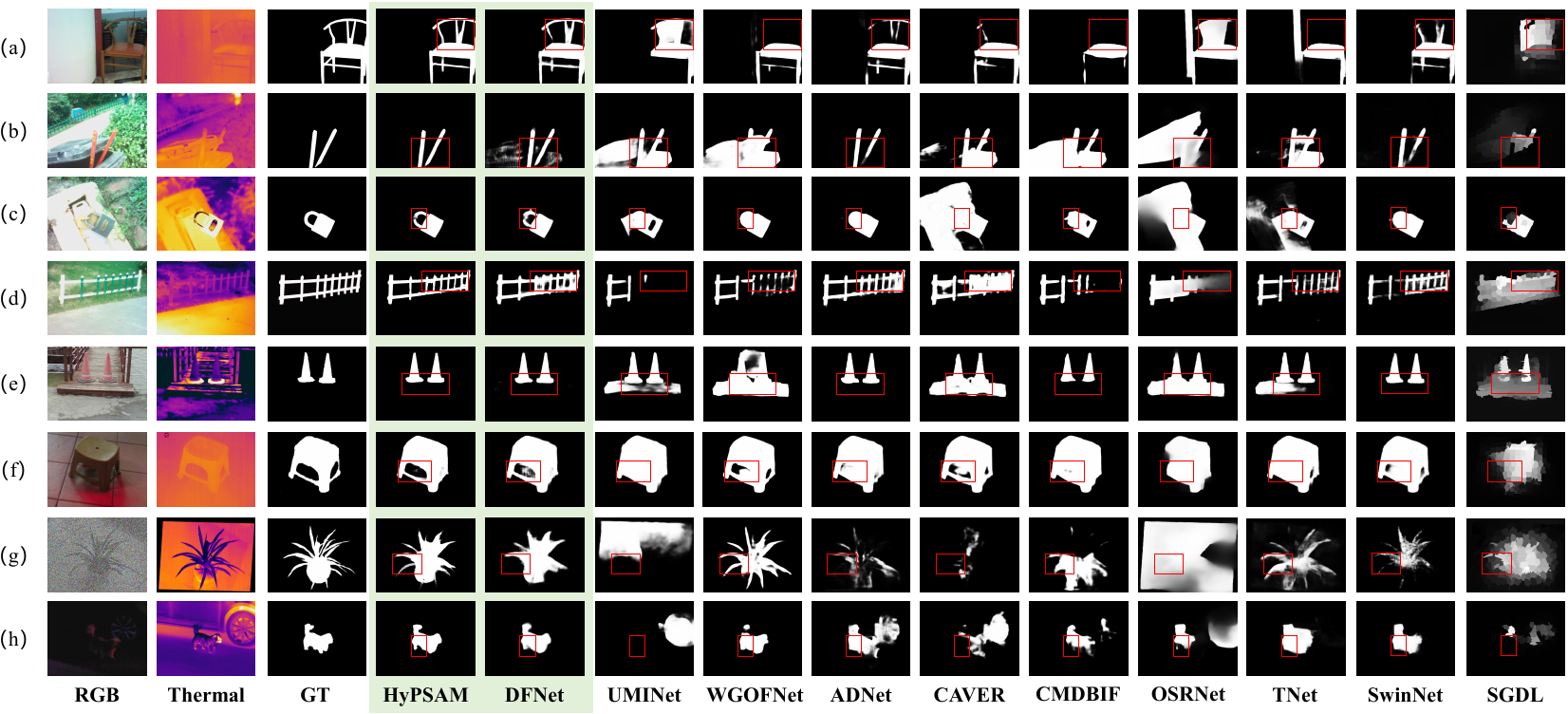}
    
    \caption{Qualitative comparison with nine state-of-the-art methods. We select eight RGB-T image pairs with diverse challenges from three datasets for saliency map comparison. From left to right, the columns are RGB images, thermal 
 images, ground truths, and the results of the ten methods. The key qualitative differences are highlighted with red bounding boxes.}
    \label{fig:vis}
\end{figure*}

\begin{table*}[t]
\setlength\tabcolsep{7pt} 
\scriptsize
  \centering
  \caption{Performance comparison (F-measure, $F_w\uparrow$) on 13 challenging attributes of the VT5000. The best and second-best results are highlighted in \textbf{bold} and \underline{underline}. $\uparrow$ $(\downarrow)$ denotes larger is better (smaller is better).
 }
 \resizebox{\textwidth}{!}{
    \begin{tabular*}{\linewidth}{@{\extracolsep{\fill}}r|c|ccccccccccccc}
    \toprule
    Attributes & Pub.Info & BSO   & CB    & CIB   & IC    & LI    & MSO   & OF    & SSO   & SA    & TC    & BW    & bRGB   & bT \\
    \midrule
    SGDL~\cite{1000}  & $\text{TMM20}$ & 0.559  & 0.522  & 0.490  & 0.482  & 0.530  & 0.513  & 0.566  & 0.565  & 0.428  & 0.473  & 0.443  & 0.568  & 0.575  \\
    CSRNet~\cite{CSRNet} &$\text{TCSVT21}$ & 0.824  & 0.767  & 0.776  & 0.746  & 0.813  & 0.760  & 0.800  & 0.718  & 0.749  & 0.753  & 0.697  & 0.798  & 0.805  \\
    CGFNet~\cite{cgfnet} & $\text{TCSVT22}$ & 0.837  & 0.825  & 0.808  & 0.792  & 0.834  & 0.786  & 0.816  & 0.768  & 0.808  & 0.806  & 0.750  & 0.836  & 0.837  \\
    SwinNet~\cite{Swinnet} & $\text{TCSVT22}$ & 0.879  & 0.835  & 0.865  & 0.820  & 0.870  & 0.819  & 0.843  & 0.738  & 0.825  & 0.827  & 0.786  & 0.846  & 0.850  \\
    ADF~\cite{5000}   & $\text{TMM22}$ & 0.785  & 0.713  & 0.749  & 0.701  & 0.739  & 0.705  & 0.711  & 0.495  & 0.677  & 0.673  & 0.683  & 0.725  & 0.732  \\
    TNet~\cite{TNet}  & $\text{TMM22}$ & 0.845  & 0.835  & 0.816  & 0.793  & 0.844  & 0.822  & 0.822  & 0.799  & 0.825  & 0.819  & 0.767  & 0.844  & 0.845  \\
    OSRNet~\cite{OSRNet} &  $\text{TIM22}$ & 0.838  & 0.801  & 0.798  & 0.764  & 0.819  & 0.782  & 0.793  & 0.707  & 0.764  & 0.769  & 0.766  & 0.810  & 0.815  \\
    ACMANet~\cite{ASY} &  $\text{KBS22}$ & 0.852  & 0.807  & 0.816  & 0.784  & 0.850  & 0.775  & 0.795  & 0.696  & 0.770  & 0.786  & 0.777  & 0.827  & 0.831  \\
    MCFNet~\cite{MCFNet} &  $\text{AI22}$ & 0.858  & 0.827  & 0.839  & 0.788  & 0.858  & 0.805  & 0.820  & 0.762  & 0.813  & 0.806  & 0.804  & 0.840  & 0.842  \\
    CMDBIF~\cite{CMDBIF-Net} &$\text{TCSVT23}$ & 0.851  & 0.833  & 0.821  & 0.803  & 0.850  & 0.800  & 0.829  & 0.816  & 0.827  & 0.827  & 0.777  & 0.851  & 0.851  \\
    CAVER~\cite{CAVER} & $\text{TIP23}$ & 0.870  & 0.844  & 0.858  & 0.810  & 0.863  & 0.817  & 0.825  & 0.780  & 0.815  & 0.827  & 0.813  & 0.854  & 0.853  \\
    ADNet~\cite{adnet} & $\text{MMA23}$ & {0.893} & {0.883} & {0.884} & {0.861} & 0.863  & {0.868} & 0.844  & {0.851} & 0.848  & {0.882} & {0.851} & {0.892} & {0.886} \\
    WGOFNet~\cite{WGOFNet} & $\text{TOMM24}$ & 0.889  & 0.865  & 0.874  & 0.837  & {0.880} & 0.845  & 0.863 & 0.844  & {0.851} & 0.849  & 0.790  & 0.877  & 0.878  \\
    UMINet~\cite{uminet} &$\text{TVC24}$ & 0.840  & 0.814  & 0.827  & 0.769  & 0.834  & 0.789  & 0.808  & 0.742  & 0.810  & 0.781  & 0.787  & 0.824  & 0.829  \\
    SACNet~\cite{wang2024alignment}& $\text{TMM24}$& 0.899 &	0.822 &	0.890 &	0.868 &	0.888 &	0.862 &	0.862 &	0.846 &	0.861 &	0.883 &	0.832&	0.891	&0.890 \\
    ConTriNet~\cite{tang2025divide} & $\text{TPAMI25}$&\underline{0.906} &	0.890 &	\underline{0.899} 	&\underline{0.874} &	0.894 &	0.864 &	\underline{0.878} &	0.863 &	\underline{0.878} &	0.885 &	0.835&	0.898&	\underline{0.898} \\

    \midrule
    \textbf{DFNet}  &  - & 0.905 & \underline{0.894} & {0.896} & {0.869} & \underline{0.900} & \underline{0.874} & {0.876} & \underline{0.871} & \underline{0.878} & \underline{0.894} & \underline{0.855} & \underline{0.900} & \underline{0.898} \\
    $\textbf{HyPSAM}$  & - & \textbf{0.920} & \textbf{0.911} & \textbf{0.911} & \textbf{0.887} & \textbf{0.917} & \textbf{0.896} & \textbf{0.891} & \textbf{0.893} & \textbf{0.895} & \textbf{0.911} & \textbf{0.871} & \textbf{0.919} & \textbf{0.916} \\
    \bottomrule
    \end{tabular*}}%
  \label{tab:2}%
  \vspace{-0.5cm}
\end{table*}

\begin{table*}[htbp]

\centering
\scriptsize
\caption{Ablation analysis of components on three datasets. The best results are highlighted in \textbf{bold}.}
\resizebox{\textwidth}{!}{
\begin{tabular}{c|l|cccc|cccc|cccc}
\toprule
\multirow{2}{*}{No.} & \multirow{2}{*}{Settings} 
& \multicolumn{4}{c|}{VT5000} 
& \multicolumn{4}{c|}{VT1000} 
& \multicolumn{4}{c}{VT821} \\
& & $F_w \uparrow$ &$\mathcal{M} \downarrow$ & $E_m \uparrow$ & $S_m \uparrow$
  & $F_w \uparrow$ & $\mathcal{M} \downarrow$ & $E_m \uparrow$ & $S_m \uparrow$
  & $F_w \uparrow$ & $\mathcal{M} \downarrow$ & $E_m \uparrow$ & $S_m \uparrow$ \\
\midrule
1 & Baseline            & 0.867 & 0.024 & 0.942 & 0.913 & 0.911 & 0.016 & 0.946 & 0.938 & 0.850 & 0.027 & 0.929 & 0.905 \\
\midrule
2 & Baseline + DIM                         & 0.894 & 0.020 & 0.955 & 0.928 & 0.929 & 0.013 & 0.957 & 0.949 & 0.877 & 0.024 & 0.936 & 0.922 \\
3 & Baseline + MDM                         & 0.888 & 0.021 & 0.950 & 0.926 & 0.926 & 0.014 & 0.950 & 0.948 & 0.871 & 0.024 & 0.931 & 0.921 \\
4 & Baseline + MDM + DFM                   & 0.888 & 0.021 & 0.951 & 0.927 & 0.926 & 0.014 & 0.951 & 0.949 & 0.879 & 0.022 & 0.937 & 0.924 \\
5 & Baseline + DIM + MDM + DFM (DFNet)                       &0.898 & 0.020 & 0.958 & 0.930
                                           & 0.930 & 0.013 & 0.955 & 0.950
                                           & 0.881 & 0.022& 0.940 & 0.926\\
\midrule
6 & SAM + random prompts                  &  0.368&  0.221&  0.591&  0.617 &  0.564&  0.139&  0.701&  0.735 &  0.462&  0.169&  0.668&  0.682 \\
\midrule
7 & DFNet + PAS + SAM-RGB                  & 0.905 & 0.022 & 0.961 & 0.928 & 0.940 & 0.014 & 0.963 & 0.952 & 0.892 & 0.023 & 0.946 & 0.927 \\
8 & DFNet + PAS + SAM-T                    & 0.884 & 0.025 & 0.958 & 0.916 & 0.922 & 0.017 & 0.960 & 0.939 & 0.852 & 0.030 & 0.934 & 0.900 \\
9 & DFNet + PAS + QMS                      & 0.906 & 0.021 & 0.962 & 0.929 & 0.940 & 0.014 & 0.963 & 0.952 & 0.895 & 0.023 & 0.948 & 0.928 \\
10 & DFNet + PAS + QMS + RS (HyPSAM)       & \textbf{0.911} & \textbf{0.019} & \textbf{0.963} & \textbf{0.931} 
                                           & \textbf{0.944} & \textbf{0.011} & \textbf{0.965} & \textbf{0.954} 
                                           & \textbf{0.903} & \textbf{0.020} & \textbf{0.948} & \textbf{0.932} \\
\bottomrule
\end{tabular}}
\label{tab:3}
\end{table*}

\begin{figure*}[t]
\setlength\tabcolsep{2.5pt}
    \centering
    \includegraphics[width=1\textwidth]{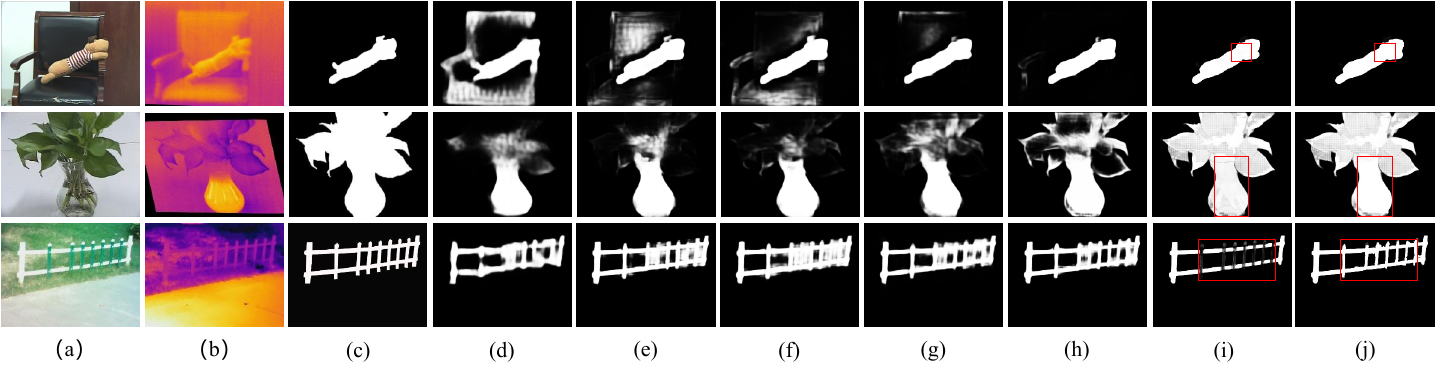}
    
    \caption{Visualization examples of ablation studies. (a) RGB image. (b) Thermal image. (c) Ground truth. (d) Baseline. (e) Baseline+DIM. (f) Baseline+MDM. (g) Baseline+MDM+DFM. (h) DFNet. (i) HyPSAM (w/o RS). (j) HyPSAM (full model).}
    \vspace{-0.5cm}
    \label{fig:10}
\end{figure*}  
\begin{figure}[t]
    \centering
    \includegraphics[width=\linewidth]{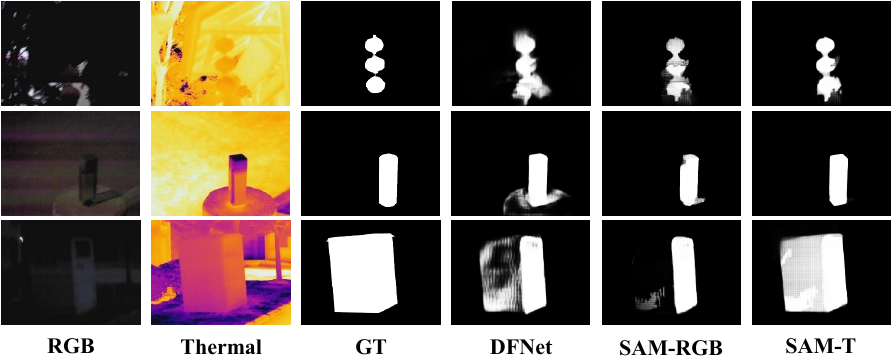}
    \caption{Comparison of SAM predictions guided by different single-modality inputs.}
    \vspace{-0.5cm}
    \label{fig:11} 
\end{figure}

\emph{2) Attribute-based Quantitative Comparison:}
To further verify the ability of different methods to handle complex scenarios, we conduct experiments on 13 challenging scenarios in the VT5000 dataset. We evaluate the $F_w$ of HyPSAM and 16 state-of-the-art methods.
The comparison results for different challenging attributes are presented in Table~\ref{tab:2}.
The results demonstrate that HyPSAM consistently outperforms competing methods across all challenging outdoor scenarios.

Specifically, compared to the second-best method, ConTriNet~\cite{tang2025divide}, our model achieves significant performance gains of 3.2\%, 2.1\%, and 1.8\% in MSO, bRGB, and bT scenarios, respectively. These gains highlight the ability of HyPSAM to resist interference, effectively mitigating the impact of challenging conditions such as low-light environments or low-quality input data.
The superior performance can be attributed to context-aware dynamic convolutions, which adaptively handle multi-modal feature interactions to provide robust saliency prompts. Additionally, by dynamically selecting the highest-quality modality and integrating hybrid prompts, HyPSAM ensures robust detection even in the presence of degraded or ambiguous inputs.

\emph{3) Qualitative Comparison:}
We present representative results of the proposed method and nine RGB-T SOD methods in various challenging scenarios, as shown in Fig.\ref{fig:vis}. 
In scenarios shown in Fig.\ref{fig:vis} (a) and (b), the object is not prominent in the thermal images, and other methods fail to effectively suppress interference from the thermal infrared modality, leading to incomplete objects, such as the back of the chair being inaccurately segmented. The inability to filter irrelevant thermal cues compromises object integrity. Figure~\ref{fig:vis} (c)–(f) depict scenes with significant challenges, where existing methods struggle to fully exploit complementary information from both modalities, leading to ambiguous object boundaries and hollow regions within salient objects. In the cases shown in Fig.\ref{fig:vis} (g) and (h), low illumination and noisy environments make it difficult to segment the pineapple and dog completely. The interference from poor-quality RGB images hinders the detection process. CAVER~\cite{CAVER} may overly rely on the RGB modality, failing to exploit thermal information under these conditions, resulting in suboptimal detection. 
Qualitative comparisons further verify that the proposed method effectively refines object details, preventing false positive detections in various challenging scenarios.

\subsection{Ablation Studies}
We investigate the importance and contributions of different modules within the proposed method on three benchmarks. The ablation studies are mainly divided into two parts: component ablation and prompt ablation.

\emph{1) Effectiveness of Components: }
Table~\ref{tab:3} shows the quantitative results of different components. The first row represents the baseline, where shallow and deep features from the backbone are simply merged via element-wise addition. The second row evaluates the impact of the dynamic interaction module (DIM), which facilitates cross-modal fusion by simultaneously exploring different dimensions' relations across modalities. The third row incorporates the multi-branch decoding module (MDM), which disentangles modality-specific and shared features to reduce the impact of modality bias, yielding further performance gains. The fourth row shows the results of applying the decision fusion module (DFM), which can adaptively fuse the cross-modal complementary information at the decision level. The fifth row demonstrates the effectiveness of DFNet, where all components work synergistically to exploit the various relations among multi-modal features at different dimensions, achieving sufficient feature fusion, visualization examples of ablation studies, as shown in Fig.\ref{fig:10}.

We also perform ablation studies on P2RNet, focusing on the quality-aware modality selector (QMS), prompt augmentation strategy (PAS), and refinement strategy (RS), with results also summarized in Table~\ref{tab:3}. As a baseline, we report the performance of SAM used independently with random prompts on RGB inputs (Row 6), aiming to isolate the effect of prompt engineering and emphasize the contribution of DFNet.
To further validate the necessity of the QMS module, we analyze the segmentation performance of SAM when guided by different single-modality inputs, as illustrated in Fig.~\ref{fig:11}. SAM performs reasonably well in well-lit scenes with RGB inputs, but its effectiveness significantly deteriorates under low-light or visually ambiguous conditions. Notably, although SAM is pretrained solely on RGB images, it can still yield meaningful segmentation results on thermal inputs. This is because SAM relies not only on color information but also on structural priors, edge cues, and learned visual semantics, which can still be partially preserved in aligned thermal images.
The ablation results clearly demonstrate that QMS enhances robustness by adaptively selecting the most informative modality, thereby avoiding noisy or misleading guidance. 
The RS can fully integrate the initial saliency map and segmentation map, ensuring that both global object structures and fine-grained details are accurately captured, leading to a notable improvement in overall performance.

\begin{table}[t]
\centering
\caption{Comparison with different prompts on the VT5000 dataset. The best results are highlighted in \textbf{bold}.}
\resizebox{\linewidth}{!}{
\scriptsize
\begin{tabular}{c|l|cccc}
\toprule
{No.} & {Prompt Type} & $F_w \uparrow$ & $\mathcal{M} \downarrow$ & $E_m \uparrow$ & $S_m \uparrow$ \\
\midrule
1 & Point                   & 0.832 & 0.063 & 0.912 & 0.875 \\
2 & Box                     & 0.903 & 0.022 & 0.957 & 0.927 \\
3 & Mask                    & 0.549 & 0.091 & 0.942 & 0.816 \\
4 & Point + Box             & 0.881 & 0.028 & 0.943 & 0.915 \\
5 & Point + Mask            & 0.823 & 0.064 & 0.919 & 0.876 \\
6 & Box + Mask   & \textbf{0.911} & \textbf{0.019} & \textbf{0.963} & \textbf{0.931} \\
7 & Point + Box + Mask      & 0.871 & 0.031 & 0.944 & 0.911 \\
\bottomrule
\end{tabular}}
\label{tab:prompt}
\end{table}

\begin{table}[t]
\centering
\caption{{Comparison with different refinement strategies on the VT5000 dataset. The best results are highlighted in \textbf{bold}.}}
\resizebox{\linewidth}{!}{
\scriptsize
\begin{tabular}{c|l|cccc}
\toprule
{No.} & {Refinement Strategies} & $F_w \uparrow$ & $\mathcal{M} \downarrow$ & $E_m \uparrow$ & $S_m \uparrow$ \\
\midrule
1 & Add                   & 0.900&  0.021&  0.959&  \textbf{0.934} \\
2 & Max                    & {0.911} &\textbf{0.019}& \textbf{0.963}& 0.931 \\
3 & CRF            & 0.879&  \textbf{0.019}&  0.929&  0.891 \\
4 & F-BRS &        0.907& \textbf{0.019}&  0.962&  0.930              \\ 
5 & Weighted Add        & 0.898&  0.021&  0.958&  0.933 \\
6 & Morphological Refinement& \textbf{0.912}&  \textbf{0.019}&  0.962&  0.926\\
\bottomrule
\end{tabular}}
\vspace{-0.5cm}
\label{tab:5}
\end{table}

\begin{figure}[t]
\setlength\tabcolsep{2.5pt}
    \centering
    \includegraphics[width=\linewidth]{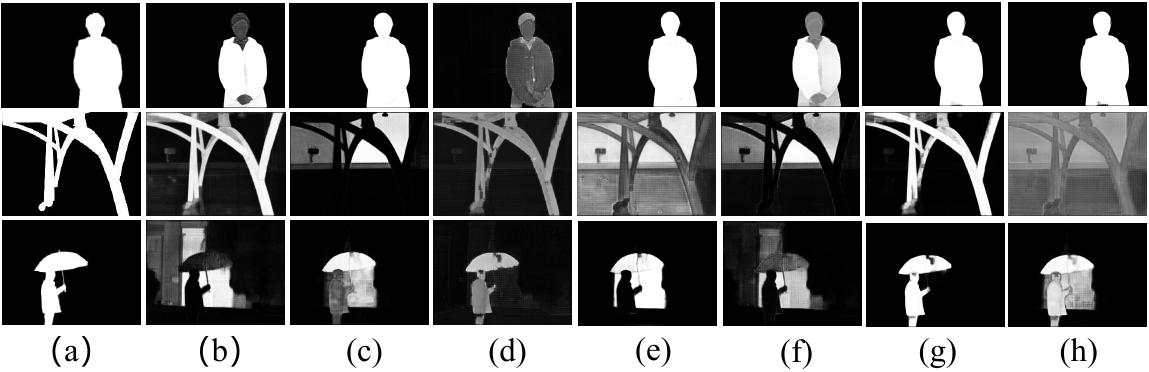}
    
    \caption{Visualization examples of different prompts. (a) Ground truth. (b) Point. (c) Box. (d) Mask. (e) Point+Box. (f) Point+Mask. (g) Box+Mask. (h) Point+Box+Mask.}
     \vspace{-0.5cm}
    \label{fig:12}
\end{figure}

\begin{figure}[t]
\setlength\tabcolsep{2.5pt}
    \centering
    \includegraphics[width=\linewidth]{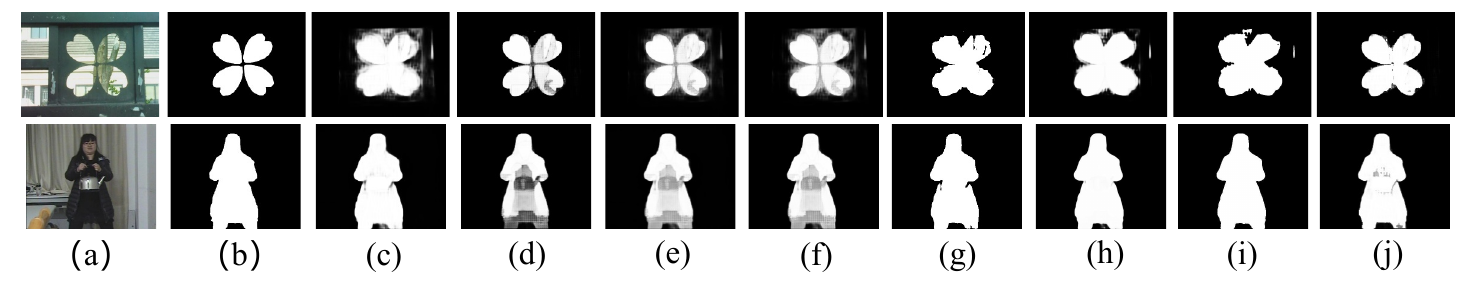}
    
    \caption{Visualization examples of different refinement strategies. (a) RGB. (b) Ground truth. (c) DFnet. (d) HyPSAM (w/o RS). (e) Add. (f) Weighted Add. (g) CRF. (h) F-BRS. (i)Morphological Refinement. (j) Max.}
     \vspace{-0.5cm}
    \label{fig:13}
\end{figure}  

\begin{table}[t]
\scriptsize
\setlength\tabcolsep{3pt}
\centering
\caption{Comparison with different methods on unaligned RGB-T datasets. The best results are highlighted in \textbf{bold}.}
\resizebox{\linewidth}{!}{
\begin{tabular}{lr|ccccccc}
\toprule
& Metric
& \makecell{SwinNet\\~\cite{Swinnet}} 
& \makecell{TNet\\~\cite{TNet}} 
& \makecell{ OSRNet\\~\cite{OSRNet}} 
& \makecell{MCFNet\\~\cite{MCFNet}} 
& \makecell{{SACNet}\\~\cite{wang2024alignment}} 
& \makecell{DFNet\\Ours} 
& \makecell{HyPSAM\\Ours} \\
\midrule

\multirow{4}{*}{\rotatebox{90}{un-VT5000}} 
& $F_w\uparrow$        & 0.823 & 0.806 & 0.571 & 0.757 & 0.876 & 0.880 & \textbf{0.895} \\
& $\mathcal{M}$$\downarrow$    & 0.031 & 0.038 & 0.106 & 0.044 & 0.023 & 0.023 & \textbf{0.022} \\
& $E_m\uparrow$        & 0.923 & 0.910 & 0.770 &0.905 & 0.949 & 0.947 & \textbf{0.955} \\
& $S_m\uparrow$        & 0.899 & 0.879 & 0.724 &0.864 & 0.911 & 0.922 & \textbf{0.922} \\
\midrule

\multirow{4}{*}{\rotatebox{90}{un-VT1000}} 
& $F_w\uparrow$        & 0.890 & 0.877 & 0.701 & 0.833 & 0.923 & 0.925 & \textbf{0.942} \\
& $\mathcal{M}$$\downarrow$    & 0.018 & 0.025 & 0.077 & 0.028 & 0.014 & 0.014 & \textbf{0.013} \\
& $E_m\uparrow$        & 0.938 & 0.927 & 0.825 & 0.929 & 0.954 & 0.950 & \textbf{0.961} \\
& $S_m\uparrow$        & 0.936 & 0.920 & 0.800 & 0.914 & 0.941 & 0.948 & \textbf{0.951} \\
\midrule

\multirow{4}{*}{\rotatebox{90}{un-VT821}} 
& $F_w\uparrow$        & 0.799 & 0.788 & 0.575 & 0.741 & 0.857 & 0.864 & \textbf{0.880} \\
& $\mathcal{M}$$\downarrow$    & 0.036 & 0.047 & 0.086 & 0.044 & 0.026 & 0.025 & \textbf{0.023} \\
& $E_m\uparrow$        & 0.905 & 0.889 & 0.790 & 0.899 & 0.929 & 0.928 & \textbf{0.936} \\
& $S_m\uparrow$        & 0.888 & 0.873 & 0.733 & 0.867 & 0.905 & 0.917 & \textbf{0.920} \\
\bottomrule
\end{tabular}}
\label{tab:6}
\end{table}

\begin{table}[t]
\scriptsize
  \centering
  \caption{Comparison with state-of-the-art methods on RGB-Nir datasets. The best results are highlighted in \textbf{bold}.}
    \begin{tabular*}{\linewidth}{@{\extracolsep{\fill}}c|cccccc}
    \toprule
    \multirow{2}[4]{*}{Methods}       &  \multicolumn{6}{c}{RGBN}    \\
    &$F_{{avg}}\uparrow$ & $F_{{max}}\uparrow$ & $F_w\uparrow$     & $\mathcal{M} \downarrow$   & $E_m\uparrow$    & $S_m\uparrow$         \\
    \midrule
RC~\cite{cheng2014global} & 0.664&  0.736&  0.442&  0.148&  0.810&  0.724 	  \\  
     DCL~\cite{li2016deep} & 0.779&  0.838&  0.660&  0.076&  0.881&  0.796  \\
     SOD8s+~\cite{song2022deep}& 0.803&  0.850&  0.745&  0.061&  0.894&  0.828 \\
    \midrule
    DFNet & 0.936&  0.956&  0.931&  0.017&  0.976&  0.946  
   \\
   HyPSAM & \textbf{0.957} &	\textbf{0.963} &	\textbf{0.946} &	\textbf{0.015} 
 & \textbf{0.978} &	\textbf{0.951} 
 \\
    \bottomrule
    \end{tabular*}%
  \label{tab:7}%
\end{table}

\begin{table}[htbp]
\scriptsize
\setlength\tabcolsep{3pt}
\centering
\caption{Comparison with state-of-the-art methods on RGB-D datasets. The best results are highlighted in \textbf{bold}.}
\resizebox{\linewidth}{!}{
\begin{tabular}{lr|ccccccc}
\toprule
 & Metric
& \makecell{MMNet\\\cite{gao2021unified}} 
& \makecell{SwinNet\\\cite{Swinnet}} 
& \makecell{CATNet\\\cite{sun2024catnet}} 
& \makecell{CPNet\\\cite{hu2024cross}} 
& \makecell{BTNet\\\cite{ren2025bio}} 
& \makecell{DFNet\\Ours} 
& \makecell{HyPSAM\\Ours} \\
\midrule

\multirow{4}{*}{\rotatebox{90}{NLPR}} 
& $F_w\uparrow$        & 0.889 & 0.908 & 0.912 & 0.918 & 0.917 & 0.916 & \textbf{0.925} \\
& $\mathcal{M}$$\downarrow$      & 0.024 & 0.018 & 0.018 & \textbf{0.016} & \textbf{0.016} & 0.017 & \textbf{0.016} \\
& $E_m\uparrow$        & 0.950 & 0.967 & 0.967 & 0.970 & 0.969 & 0.970 & \textbf{0.971} \\
& $S_m\uparrow$        & 0.250 & 0.941 & 0.940 & 0.940 & 0.941 & \textbf{0.944} & \textbf{0.944} \\
\midrule

\multirow{4}{*}{\rotatebox{90}{NJU2K}} 
& $F_w\uparrow$        & 0.900 & 0.922 & 0.919 & 0.923 & 0.917 & \textbf{0.927} & \textbf{0.931} \\
& $\mathcal{M}$$\downarrow$      & 0.038 & 0.027 & 0.026 & 0.025 & 0.025 & \textbf{0.022} & 0.023 \\
& $E_m\uparrow$        & 0.919 & 0.934 & 0.933 & 0.935 & 0.932 & 0.939 & \textbf{0.942} \\
& $S_m\uparrow$        & 0.911 & 0.935 & 0.932 & 0.935 & 0.932 & \textbf{0.940} & \textbf{0.940} \\
\midrule
\multirow{4}{*}{\rotatebox{90}{STERE}} 
& $F_w\uparrow$        & 0.880 & 0.893 & 0.894 & 0.895 & 0.902 & 0.899 & \textbf{0.915} \\
& $\mathcal{M}$$\downarrow$    & 0.045 & 0.033 & 0.030 & 0.029 & 0.028 & 0.029 & \textbf{0.026} \\
& $E_m\uparrow$        & 0.924 & 0.929 & 0.936 & 0.933 & 0.940 & 0.936 & \textbf{0.945} \\
& $S_m\uparrow$        & 0.891 & 0.919 & 0.921 & 0.920 & 0.927 & 0.929 & \textbf{0.934} \\

\bottomrule
\end{tabular}}
\label{tab:8}
\end{table}

\emph{2) Effectiveness of Prompts:}
We conduct prompt ablation studies to further investigate the impact of different prompts, as summarized in Table~\ref{tab:prompt}. We first evaluate three types of individual prompts: points, boxes, and masks, where the point represents the centroids of the mask. Their results are presented in the first three rows of Table~\ref{tab:prompt}. Visualization examples of different prompts as shown in Fig.\ref{fig:12}. It can be observed that the box prompt yields superior performance due to its ability to provide global structured spatial information for guiding segmentation effectively. In contrast, point prompts are too sparse, offering limited contextual information, which leads to suboptimal results. Surprisingly, the mask prompt performs the worst. When used alone, it often causes SAM to generate soft probability maps with background noise and even over-segment the object. The mask serves as a dense and strict prior, making SAM closely follow it. As a result, any noise or inaccurate edges in the input mask are retained or even amplified, leading to poorer segmentation quality.

We then explore the combination of these prompts to investigate their complementary effects. The combination of all prompts does not yield competitive results. This may be due to conflicting information introduced by point prompts, which can create inconsistencies when combined with richer spatial information from boxes and masks. Instead, the combination of box and mask prompts achieves the best performance, as the box defines clear regions of interest, while the mask refines object boundaries, thereby striking an effective balance between global and local guidance.

\emph{3) Effectiveness of Refinement Strategies:}
Table~\ref{tab:5} compares six refinement strategies on the VT5000 dataset, including add, max, CRF, F-BRS~\cite{sofiiuk2020f}, entropy-based weighted fusion, and morphological refinement, with visual examples shown in Fig.~\ref{fig:13}. The advantage of max fusion lies in its ability to retain the most confident activations from both the initial saliency map and the SAM-predicted mask. Since SAM often produces fine-grained results that split semantically unified objects (\emph{e.g.}, a person holding a kettle) into separate instances, additive or weighted fusion may amplify redundant or ambiguous boundaries. While F-BRS and morphological refinement achieve performance comparable to max fusion, the latter is significantly more efficient and simpler to implement. Therefore, we adopt max fusion as the default refinement strategy in our framework.

\begin{table*}[htbp]
\setlength\tabcolsep{1.8pt}
\scriptsize
  \centering
  \caption{The generalization of HyPSAM is tested on three benchmarks, encompassing three distinct architectural paradigms. The best results are highlighted in \textbf{bold}. ${\dagger}$ indicates the results are refined by HyPSAM.}
  \resizebox{\linewidth}{!}{
    \begin{tabular}{c|c|cccccc|cccccc|cccccc}
    \toprule
    
   \multirow{2}[2]{*}{Types} & \multicolumn{1}{c|}{\multirow{2}[2]{*}{Methods}} & \multicolumn{6}{c|}{VT5000}                   & \multicolumn{6}{c|}{VT1000}                   & \multicolumn{6}{c}{VT821} \\
          &       &  $F_{avg}\uparrow$  & $F_{max}\uparrow$  & $F_w\uparrow$    & $\mathcal{M} \downarrow$   & $E_m\uparrow$    & $S_m\uparrow$   & $F_{avg}\uparrow$  & $F_{max}\uparrow$  & $F_w\uparrow$    & $\mathcal{M} \downarrow$   & $E_m\uparrow$    & $S_m\uparrow$  & $F_{avg}\uparrow$  & $F_{max}\uparrow$  & $F_w\uparrow$    & $\mathcal{M} \downarrow$   & $E_m\uparrow$    & $S_m\uparrow$  \\ 
    \midrule
   \multirow{2}[2]{*}{\makecell{Single-\\stream}}  & OSRNet~\cite{OSRNet} & 0.823  & 0.866  & 0.807  & 0.040  & 0.908  & 0.875  & 0.892  & 0.929  & 0.891  & 0.022  & 0.935  & 0.926  & 0.814  & 0.862  & 0.801  & 0.043  & 0.896  & 0.875  \\
          & $\text{OSRNet}^{\dagger}$ & \textbf{0.857} & \textbf{0.874} & \textbf{0.829} & \textbf{0.037} & \textbf{0.918 } & \textbf{0.882} & \textbf{0.919} & \textbf{0.932} & \textbf{0.910} & \textbf{0.020} & \textbf{0.947} & \textbf{0.932} & \textbf{0.849} & \textbf{0.867} & \textbf{0.826} & \textbf{0.036} & \textbf{0.912} & \textbf{0.882} \\
    \midrule
    \multirow{30}[1]{*}{\makecell{Dual-\\stream}} 
    
    & CSRNet~\cite{CSRNet} & 0.811  & 0.857  & 0.796  & 0.042  & 0.905  & 0.868  & 0.877  & \textbf{0.918 } & 0.878  & 0.024  & 0.925  & 0.918  & 0.831  & 0.880  & 0.821  & 0.038  & 0.909  & 0.885  \\
          &  $\text{CSRNet}^{\dagger}$ & \textbf{0.843} & \textbf{0.863} & \textbf{0.814} & \textbf{0.039} & \textbf{0.914} & \textbf{0.872} & \textbf{0.903} & \textbf{0.918} & \textbf{0.893} & \textbf{0.022} & \textbf{0.937} & \textbf{0.924} & \textbf{0.862} & \textbf{0.882} & \textbf{0.843} & \textbf{0.032} & \textbf{0.920} & \textbf{0.892} \\
          \cmidrule{2-20}       
          & CGFNet~\cite{cgfnet} & 0.851  & 0.887  & 0.831  & 0.035  & 0.922  & 0.883  & 0.906  & 0.936  & 0.900  & 0.023  & 0.944  & 0.923  & 0.845  & 0.885  & 0.829  & 0.038  & 0.912  & 0.881  \\
          &  $\text{CGFNet}^{\dagger}$ & \textbf{0.879} & \textbf{0.895} & \textbf{0.847} & \textbf{0.033} & \textbf{0.927} & \textbf{0.889} & \textbf{0.926} & \textbf{0.938} & \textbf{0.913} & \textbf{0.021} & \textbf{0.950} & \textbf{0.930} & \textbf{0.876} & \textbf{0.893} & \textbf{0.847} & \textbf{0.035} & \textbf{0.919} & \textbf{0.889} \\
          \cmidrule{2-20}   
          & SwinNet~\cite{Swinnet} & 0.865  & 0.915  & 0.846  & 0.026  & 0.942  & 0.912  & 0.896  & 0.948  & 0.894  & 0.018  & 0.947  & 0.938  & 0.847  & 0.903  & 0.818  & 0.030  & 0.926  & 0.904  \\
          & $\text{SwinNet}^{\dagger}$ & \textbf{0.910} & \textbf{0.924} & \textbf{0.888} & \textbf{0.023} & \textbf{0.950} & \textbf{0.918} & \textbf{0.940} & \textbf{0.952} & \textbf{0.935} & \textbf{0.013} & \textbf{0.961} & \textbf{0.948} & \textbf{0.895} & \textbf{0.915} & \textbf{0.876} & \textbf{0.024} & \textbf{0.934} & \textbf{0.914} \\
          \cmidrule{2-20}   
          & ADF~\cite{5000}   & 0.778  & 0.863  & 0.722  & 0.048  & 0.891  & 0.864  & 0.847  & 0.923  & 0.804  & 0.034  & 0.921  & 0.910  & 0.717  & \textbf{0.804} & 0.627  & 0.077  & \textbf{0.843} & 0.810  \\
          & $\text{ADF}^{\dagger}$  & \textbf{0.850} & \textbf{0.871} & \textbf{0.812} & \textbf{0.038} & \textbf{0.909} & \textbf{0.873} & \textbf{0.914} & \textbf{0.930} & \textbf{0.898} & \textbf{0.024} & \textbf{0.940} & \textbf{0.925} & \textbf{0.766} & 0.798  & \textbf{0.726} & \textbf{0.058} & 0.835  & \textbf{0.822} \\
          \cmidrule{2-20}   
          & TNet~\cite{TNet}  & 0.846  & 0.895  & 0.840  & 0.033  & 0.927  & 0.895  & 0.889  & 0.937  & 0.895  & 0.021  & 0.937  & 0.929  & 0.842  & 0.904  & 0.841  & 0.030  & 0.919  & 0.899  \\
          & $\text{TNet}^{\dagger}$ & \textbf{0.885} & \textbf{0.901} & \textbf{0.860} & \textbf{0.031} & \textbf{0.937} & \textbf{0.901} & \textbf{0.922} & \textbf{0.939} & \textbf{0.914} & \textbf{0.019} & \textbf{0.952} & \textbf{0.935} & \textbf{0.889} & \textbf{0.908} & \textbf{0.872} & \textbf{0.027} & \textbf{0.932} & \textbf{0.911} \\
          \cmidrule{2-20}   
          & ACMANet~\cite{ASY} & 0.858  & 0.890  & 0.823  & 0.033  & \textbf{0.932} & 0.887  & 0.904  & 0.933  & 0.889  & 0.021  & 0.945  & 0.927  & 0.837  & 0.873  & 0.807  & 0.035  & 0.914  & 0.883  \\
          & $\text{ACMANet}^{\dagger}$  & \textbf{0.878} & \textbf{0.895} & \textbf{0.854} & \textbf{0.030} & \textbf{0.932} & \textbf{0.895} & \textbf{0.921} & \textbf{0.934} & \textbf{0.917} & \textbf{0.018} & \textbf{0.948} & \textbf{0.935} & \textbf{0.856} & \textbf{0.876} & \textbf{0.839} & \textbf{0.029} & \textbf{0.918} & \textbf{0.890} \\
          \cmidrule{2-20}   
          & MCFNet~\cite{MCFNet} & 0.848  & \textbf{0.886} & 0.836  & 0.033  & 0.924  & 0.887  & 0.902  & \textbf{0.939} & 0.906  & 0.019  & 0.944  & 0.932  & 0.844  & \textbf{0.889 } & 0.835  & 0.029  & \textbf{0.918} & 0.891  \\
          & $\text{MCFNet}^{\dagger}$ & \textbf{0.868} & 0.884  & \textbf{0.846} & \textbf{0.032} & \textbf{0.927} & \textbf{0.892} & \textbf{0.926} & 0.938  & \textbf{0.921 } & \textbf{0.016} & \textbf{0.952} & \textbf{0.938} & \textbf{0.863} & 0.884  & \textbf{0.846} & \textbf{0.028 } & \textbf{0.918} & \textbf{0.896} \\
          \cmidrule{2-20}   
          & CAVER~\cite{CAVER} & 0.856  & 0.897  & 0.849  & 0.028  & 0.935  & 0.899  & 0.906  & 0.945  & 0.912  & \textbf{0.016} & 0.949  & 0.938  & 0.854  & 0.897  & 0.846  & \textbf{0.026} & 0.928  & 0.897  \\
          & $\text{CAVER}^{\dagger}$ & \textbf{0.884 } & \textbf{0.903} & \textbf{0.866} & \textbf{0.029} & \textbf{0.939} & \textbf{0.906} & \textbf{0.929} & \textbf{0.946} & \textbf{0.926} & \textbf{0.016} & \textbf{0.955} & \textbf{0.943} & \textbf{0.881} & \textbf{0.899} & \textbf{0.866} & \textbf{0.026} & \textbf{0.929} & \textbf{0.908} \\
          \cmidrule{2-20}   
          & ADNet~\cite{adnet} & 0.893  & 0.924  & 0.884  & 0.022  & 0.953  & 0.922  & 0.916  & \textbf{0.952} & 0.920  & 0.015  & 0.952  & 0.944  & 0.869  & 0.915  & 0.860  & 0.024  & 0.930  & 0.915  \\
          & $\text{ADNet}^{\dagger}$  & \textbf{0.918} & \textbf{0.930} & \textbf{0.899} & \textbf{0.021} & \textbf{0.958} & \textbf{0.923} & \textbf{0.938} & 0.950  & \textbf{0.934} & \textbf{0.014} & \textbf{0.960} & \textbf{0.947} & \textbf{0.901} & \textbf{0.919} & \textbf{0.888} & \textbf{0.021} & \textbf{0.941} & \textbf{0.922} \\
          \cmidrule{2-20}   
          & WGOFNet~\cite{WGOFNet} & 0.883  & 0.912  & 0.873  & 0.025  & 0.945  & 0.911  & 0.919  & \textbf{0.946} & 0.922  & 0.016  & 0.951  & 0.940  & 0.875  & 0.911  & 0.868  & 0.025  & 0.934  & 0.908  \\
          & $\text{WGOFNet}^{\dagger}$ & \textbf{0.900 } & \textbf{0.913} & \textbf{0.883} & \textbf{0.024} & \textbf{0.948} & \textbf{0.914} & \textbf{0.933} & 0.944  & \textbf{0.930} & \textbf{0.014} & \textbf{0.956} & \textbf{0.944} & \textbf{0.896} & \textbf{0.912} & \textbf{0.883} & \textbf{0.023} & \textbf{0.939} & \textbf{0.916} \\
          \cmidrule{2-20}   
          & UMINet~\cite{uminet} & 0.831  & 0.877  & 0.820  & 0.035  & 0.919  & 0.882  & 0.892  & 0.935  & 0.896  & 0.021  & 0.941  & 0.926  & 0.791  & 0.849  & 0.782  & \textbf{0.054} & 0.879  & 0.858  \\
          & $\text{UMINet}^{\dagger}$ & \textbf{0.864 } & \textbf{0.882} & \textbf{0.839} & \textbf{0.034} & \textbf{0.926} & \textbf{0.892} & \textbf{0.922} & \textbf{0.937} & \textbf{0.916} & \textbf{0.018} & \textbf{0.951} & \textbf{0.936} & \textbf{0.827} & \textbf{0.853} & \textbf{0.804} & \textbf{0.054} & \textbf{0.887} & \textbf{0.868} \\
          \midrule
    \multirow{4}[1]{*}{\makecell{Triple-\\stream}} & CMDBIF~\cite{CMDBIF-Net} & 0.868  & 0.892  & 0.846  & 0.032  & 0.933  & 0.886  & 0.914  & \textbf{0.931} & 0.909  & 0.019  & 0.952  & 0.927  & 0.856  & 0.887  & 0.837  & 0.032  & 0.923  & 0.882  \\
          &$\text{CMDBIF}^{\dagger}$ & \textbf{0.882} & \textbf{0.896} & \textbf{0.857} & \textbf{0.031} & \textbf{0.936} & \textbf{0.895} & \textbf{0.924} & \textbf{0.931} & \textbf{0.918} & \textbf{0.017} & \textbf{0.953} & \textbf{0.936} & \textbf{0.871} & \textbf{0.889} & \textbf{0.850} & \textbf{0.031} & \textbf{0.924} & \textbf{0.893} \\
          \cmidrule{2-20}
          &ConTriNet~\cite{tang2025divide} & 0.898 & 0.927 & 0.895 & {0.020} & 0.956 & 0.923 & 0.917 & 0.943 & 0.923 & 0.015 & 0.953 & 0.941 & 0.878 & 0.914 & 0.875 & 0.022 & 0.940 & 0.916 \\
           & $\text{ConTriNet}^{\dagger}$ & \textbf{0.918}&  \textbf{0.928}&  \textbf{0.904}&  \textbf{0.020}&  \textbf{0.959}&  \textbf{0.926}&\textbf{0.934}&  \textbf{0.945}&  \textbf{0.933}&  \textbf{0.014}&  \textbf{0.957}&  \textbf{0.945}&\textbf{0.899}&  \textbf{0.916}&  \textbf{0.890}&  \textbf{0.020}&  \textbf{0.944}&  \textbf{0.923}\\
    \bottomrule
    \end{tabular}}%
  \label{tab:9}%
\end{table*}%

\begin{table*}[htbp]
\centering
\setlength\tabcolsep{2.5pt}
\scriptsize
\caption{Performance comparison of SAMv2 and ETAM in our framework on the RGB-T dataset.}
\resizebox{\linewidth}{!}{
\begin{tabular}{c|cccccc|cccccc|cccccc}
    \toprule
    \multirow{2}[6]{*}{Models}       &  \multicolumn{6}{c|}{VT5000}&  \multicolumn{6}{c|}{VT1000}&  \multicolumn{6}{c}{VT821}    \\
     &$F_{{avg}}\uparrow$ & $F_{{max}}\uparrow$ & $F_w\uparrow$     & $\mathcal{M} \downarrow$   & $E_m\uparrow$    & $S_m\uparrow$  
      &$F_{{avg}}\uparrow$ & $F_{{max}}\uparrow$& $F_w\uparrow$     & $\mathcal{M} \downarrow$   & $E_m\uparrow$    & $S_m\uparrow$ & 
       $F_{{avg}}\uparrow$ & $F_{{max}}\uparrow$&$F_w\uparrow$     & $\mathcal{M} \downarrow$   & $E_m\uparrow$    & $S_m\uparrow$       \\
    \midrule
HyPSAM (SAMv2) & 0.928 &0.939 &0.911 &0.019 &0.963& 0.931 &0.946& 0.957& 0.944& 0.011& 0.965& 0.954&0.914 &0.930& 0.903& 0.020& 0.948& 0.932   \\
HyPSAM (ETAM)  &0.927&  0.934&  0.916&  0.018&  0.964&  0.929& 0.946&  0.949&  0.944&  0.011&  0.966&  0.950&0.911&  0.921&  0.902&  0.019&  0.950&  0.926 \\
\bottomrule
\end{tabular}}
\label{tab:10}
\end{table*}

\subsection{Application to Other SOD Tasks}
To evaluate the generalization capability of HyPSAM, we extend its application to various salient object detection tasks across unaligned RGB-T, RGB-D, and RGB-NIR modalities. HyPSAM achieves consistently strong performance.
On unaligned RGB-T datasets ~\cite{wang2024alignment} (Table~\ref{tab:6}), our method outperforms recent methods across all evaluation metrics, demonstrating its robustness to cross-modal misalignment.
For RGB-D datasets~\cite{niu2012leveraging,ju2014depth,peng2014rgbd}     (Table~\ref{tab:7}), HyPSAM achieves competitive or superior performance compared to several state-of-the-art models, including Transformer-based approaches, highlighting its semantic transferability and effective depth adaptation.
On the RGB-NIR dataset~\cite{song2022deep} (Table~\ref{tab:8}), HyPSAM attains the best results with an $F_w$ of 0.946 and a $\mathcal{M}$ of 0.015, further validating its ability to generalize across unseen modalities.

\subsection{Generalization Performance of HyPSAM}

To evaluate the generalization capability of our framework, we apply it to three types of RGB-T SOD models by generating hybrid prompts from their saliency maps. As shown in Table~\ref{tab:9}, where “${\dagger}$” denotes results after optimization, most metrics across all methods improve substantially. In addition, we observe a clear positive correlation between the accuracy of the initial saliency maps and the performance of the optimized models. In other words, the more accurate the initial maps are, the greater the improvements after optimization. These results demonstrate the effectiveness of the hybrid prompt-driven approach in refining segmentation outcomes.

\subsection{Generalization with Alternative Segmentation Models}
To further validate the generality of our proposed HyPSAM framework, we replace the SAMv2 in P2RNet with the efficient track anything model (ETAM) \cite{xiong2024efficient}, a lightweight segmentation model. Table~\ref{tab:10} reports the performance comparison. While ETAM achieves reasonable accuracy, SAMv2 yields better overall results. These results confirm that our framework is compatible with different segmentation backbones and can benefit from stronger base models.

\subsection{Complexity Analysis}
Model complexity is typically measured by parameters, FLOPs, and FPS. As shown in Table~\ref{tab:11}, 
DFNet strikes a favorable balance between performance and complexity. Each ODConv module is lightweight and computationally efficient, requiring only 0.64 ms per image with 2.385M FLOPs and 25.8K parameters. The full version of HyPSAM (SAMv2), which incorporates the original SAM with a ViT-H backbone, comprises 817.6M parameters and 3033.4 GFLOPs. Despite the high computational cost, the significant performance gains in segmentation justify the overhead. To enhance efficiency, we also introduce a lightweight variant, HyPSAM (ETAM), which reduces the parameter count to 208.6M and FLOPs to 394.3G, while maintaining competitive accuracy and achieving a real-time inference speed of 25 FPS.
\begin{table}[htbp]
\centering
\caption{Comparison of the complexity of some recent publicly available state-of-the-art RGB-T SOD methods.}
\setlength{\tabcolsep}{4pt}
\renewcommand{\arraystretch}{1.1}
\scriptsize
\begin{tabular}{r|c|c|c|c}
\toprule
\textbf{Method} & \textbf{Backbone} & \textbf{FLOPs (G)} & \textbf{Params. (M)} & FPS\\
\midrule
ADF~\cite{5000}              & VGG-16           & 128.2   & 83.1& 7 \\
MIDD~\cite{MIDD}            & VGG-16          & 216.6   & 52.4 &22\\
CGFNet~\cite{cgfnet}        & VGG-16            & 345.1   & 66.4&13 \\
OSRNet~\cite{OSRNet}        & VGG-16            & 34.3    & 15.6&53 \\
CAVER~\cite{CAVER}          & ResNet-50d              & 44.4    & 55.8 &27\\
SwinNet~\cite{Swinnet}      & Swin-B                 & 124.3   & 198.7&10 \\
 WGOFNet~\cite{WGOFNet} & PVT & 48.5   & 61.8&11  \\
  ADNet~\cite{adnet} & Swin-B+MobileViT & 56.7   &  93.2&43   \\
\midrule
DFNet & Swin-B & 74.1   &  133.3& 41\\
HyPSAM (SAMv2) & Swin-B+ViT-H& 3033.4   &  817.6&  3\\
HyPSAM (ETAM) & Swin-B+EfficientViT& 394.3   &  208.6 & 25\\
\bottomrule
\end{tabular}
\label{tab:11}
\end{table}

\subsection{Failure Cases and Analysis}
\begin{figure}[htpb]
    \centering
    \includegraphics[scale=0.5]{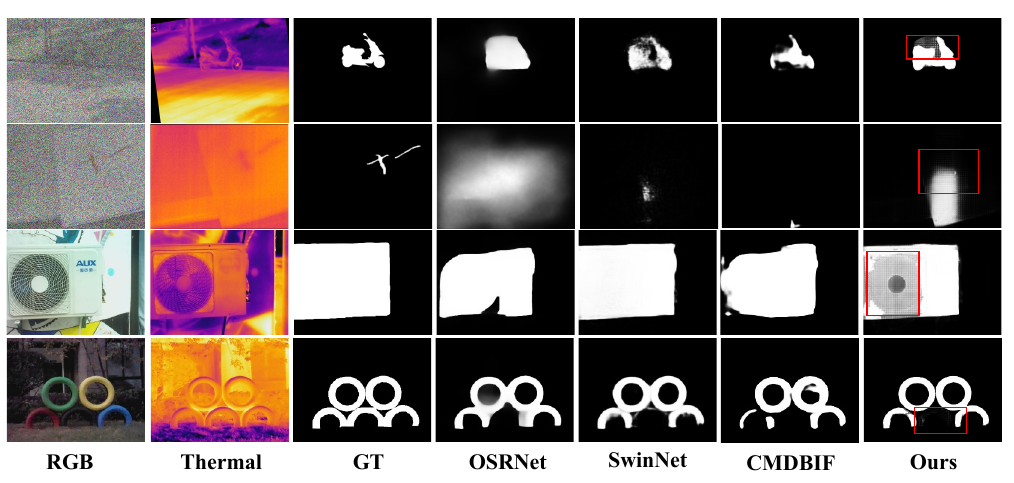}
    \caption{Visual results of our HyPSAM and other advanced methods in some typical failure cases.}
     \vspace{-0.5cm}
    \label{fig:14} 
\end{figure}

To better analyze the limitations of HyPSAM, we present failure cases in Fig.~\ref{fig:14}, compared with existing methods including OSRNet\cite{OSRNet}, SwinNet\cite{Swinnet}, and CMDBIF\cite{CMDBIF-Net}. 
The first two rows show that although the QMS correctly identifies the more reliable modality, SAM still fails in thermal inputs with low target-background contrast, producing artifacts or even incorrect object recognition.
In the third row, HyPSAM tends to over-segment semantically unified objects due to its fine-grained instance sensitivity, resulting in unnecessary internal separation. In the last row, the foreground object exhibits low contrast with the background, making it difficult for HyPSAM to distinguish foreground boundaries accurately.
These cases demonstrate that while HyPSAM significantly outperforms prior methods in structure preservation and modality adaptation, certain limitations persist in scenarios with subtle foreground-background variations and complex object details. Moreover, since SAM is pre-trained exclusively on large-scale RGB data, a domain gap remains when generalizing to challenging thermal scenarios, particularly under weak contrast. Future work may focus on enhancing prompt semantics and refining instance integration to further improve robustness in complex scenarios.

\section{Conclusion}
In this paper, we proposed a HyPSAM, a novel hybrid prompt-driven segment anything model for RGB-T SOD, which consists of key components: DFNet and P2RNet. To address the intrinsic
insufficient feature fusion, DFNet incorporated a dynamic fusion module coupled with a multi-branch decoding module to facilitate adaptive cross-modality interaction and generate robust saliency maps. To mitigate the extrinsic limitations of data scarcity, P2RNet leveraged a comprehensive hybrid prompting strategy that synergistically combines texts, boxes, and masks to drive SAM to accurately segment the saliency object. The initial saliency map and the segmentation map were then fused to produce the refined result with complete structures and clear boundaries. Extensive experiments demonstrated the superiority of our proposed method on multi-modal benchmarks. Furthermore, HyPSAM exhibits strong versatility and generalization by seamlessly integrating with different methods, achieving substantial performance gains without additional training.

\bibliographystyle{IEEEbib}
\bibliography{my_full_bib}


 
\vspace{11pt}

\vspace{-33pt}
\begin{IEEEbiography}[{\includegraphics[width=1in,height=1.25in,clip,keepaspectratio]{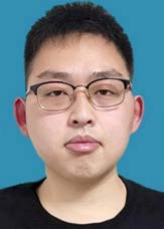}}]{Ruichao Hou} (Member, IEEE) received his Ph.D degree from the Department of Computer Science and Technology, Nanjing University in 2023.
He is currently an Assistant Researcher at the Software Institute of Nanjing University. His research mainly focuses on multi-modal object detection and tracking.
\end{IEEEbiography}

 \vspace{-33pt}
 \begin{IEEEbiography}[{\includegraphics[width=1in,height=1.25in,clip,keepaspectratio]{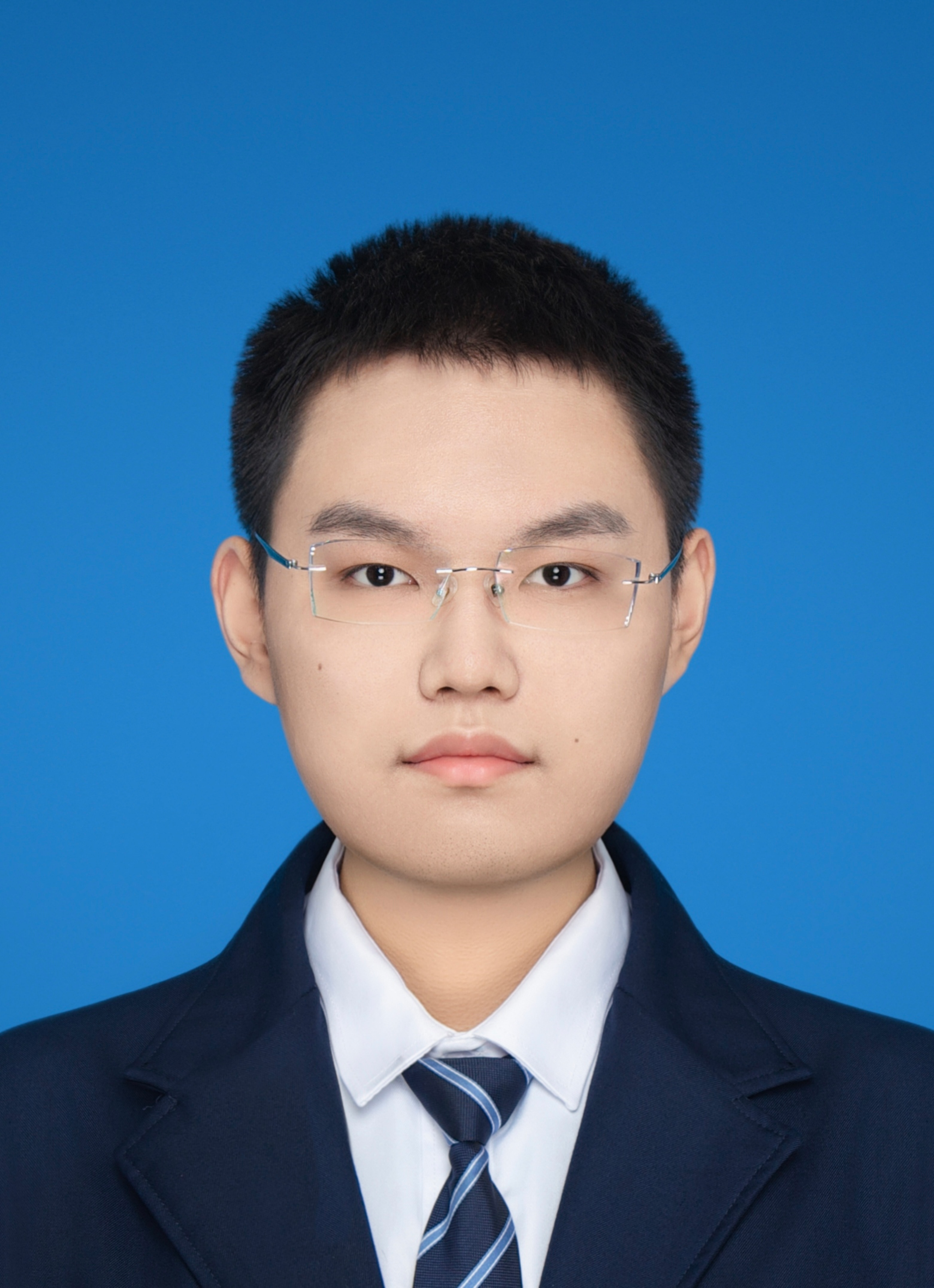}}]{Xingyuan Li} received the B.S. degree from the School of Artificial Intelligence at Nanjing University, Nanjing, China, where he is currently pursuing the Ph.D. degree at the School of Artificial Intelligence at Nanjing University. His current research interests include RGB-T salient object detection.
 \end{IEEEbiography}

 \vspace{-33pt}
 \begin{IEEEbiography}[{\includegraphics[width=1in,height=1.25in,clip,keepaspectratio]{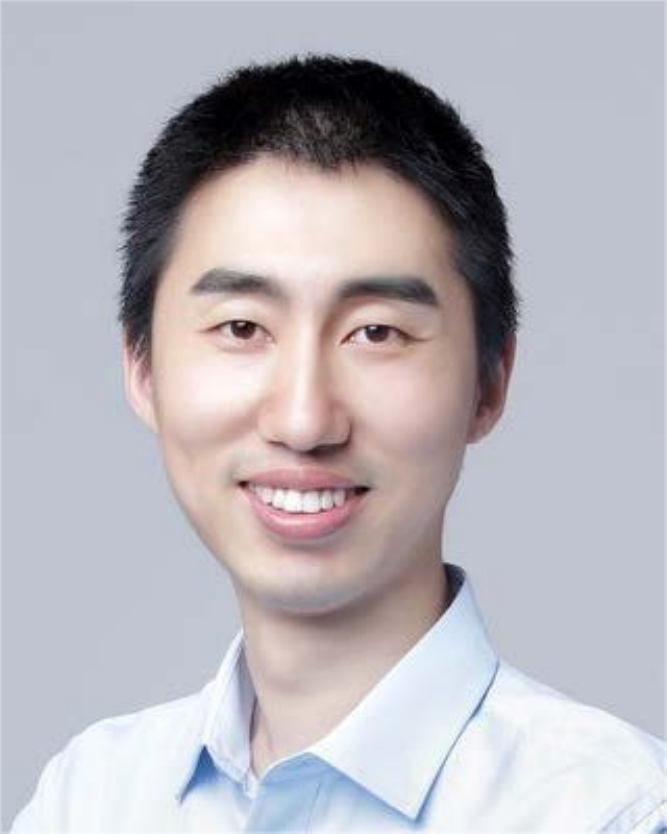}}] {Tongwei Ren} (Member, IEEE) received the B.S., M.E., and Ph.D. degrees from Nanjing University, Nanjing, China, in 2004, 2006, and 2010, respectively. He joined Nanjing University in 2010, and at present he is a professor. His research interests mainly include multimedia computing and its real-world applications. He has published more than 40 papers in top-tier journals and conferences. He was a recipient of the best paper candidate awards of ICIMCS 2014, PCM 2015, and MMAsia 2020, and he was in the champion teams of ECCV 2018 PIC challenge, MM 2019 VRU challenge, MM 2020 DVU challenge, MM 2022 DVU challenge and MM 2023 DVU challenge.
 \end{IEEEbiography}

 \vspace{-33pt}
 \begin{IEEEbiography}[{\includegraphics[width=1in,height=1.25in,clip,keepaspectratio]{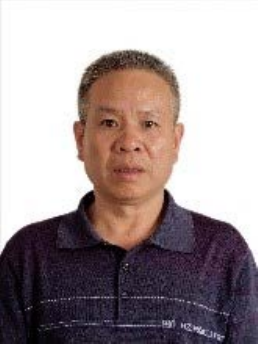}}] {Dongming Zhou} received the B.S. and M.S. degrees in industry automatization from the Department of Automatic Control Engineering, Huazhong University of Science and Technology, Wuhan, China, in 1985 and 1988, respectively, and the Ph. D. degree in circuitry
and system from the School of Information Science and Technology, Fudan University, Shanghai, China, in 2004. In 2008, he was a Visiting Scholar with York University, Toronto, ON, Canada. He is currently a Professor at the School of Information Science and Engineering, Yunnan University, Kunming, China. His current research interests include biomedical engineering, computational intelligence, complex systems, neural networks, and their applications.
 \end{IEEEbiography}

 \vspace{-33pt}
 \begin{IEEEbiography}[{\includegraphics[width=1in,height=1.25in,clip,keepaspectratio]{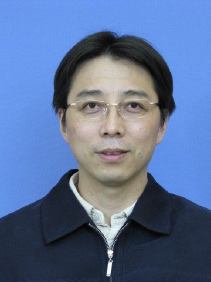}}] {Gangshan Wu} (Member, IEEE) received the B.Sc., M.S., and Ph.D. degrees from the Department of Computer Science and Technology, Nanjing University, Nanjing, China, in 1988, 1991, and 2000, respectively. He is currently a Professor with the School of Computer Science, Nanjing University. His current research interests include computer vision, multimedia content analysis, multimedia information retrieval, digital museum, and large-scale volumetric data processing.
 \end{IEEEbiography}

\vspace{-33pt}
\begin{IEEEbiography}[{\includegraphics[width=1in,height=1.25in,clip,keepaspectratio]{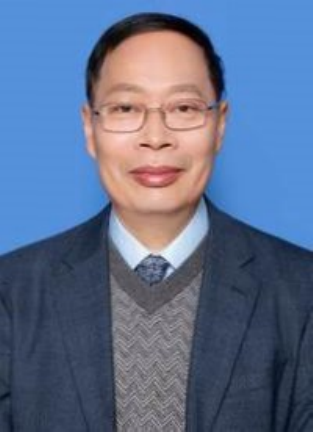}}]{Jinde Cao}(Fellow, IEEE) received the B.S. degree in mathematics/applied mathematics from Anhui Normal University, Wuhu, China, in 1986, the M.S. degree in mathematics/applied mathematics from Yunnan University, Kunming, China, in 1989, and the Ph.D. degree in mathematics/applied mathematics from Sichuan University, Chengdu, China, in 1998. He is currently an Endowed Chair Professor, the
Dean of the School of Mathematics, the Director of the Jiangsu Provincial Key Laboratory of Networked Collective Intelligence of China and the Research Center for Complex Systems and Network Sciences, Southeast University, Nanjing, China.

Prof. Cao is elected as a member of the Academy of Europe and the European Academy of Sciences and Arts, a Foreign Member of the Russian
Academy of Natural Sciences and the Lithuanian Academy of Sciences, a fellow of the Pakistan Academy of Sciences and the African Academy of
Sciences, and an IASCYS Academician. He was a recipient of the National Innovation Award of China, the Obada Prize, and the Highly Cited Researcher Award in Engineering, Computer Science, and Mathematics by Thomson Reuters/Clarivate Analytics.
\end{IEEEbiography}


\end{document}